\documentclass[a4paper,10pt]{article}

\usepackage[english]{babel}
\usepackage{xspace}
\usepackage{titling}
\usepackage[verbose,a4paper,lmargin=3cm,rmargin=3cm,tmargin=2.5cm,bmargin=2.5cm]{geometry}

\usepackage[dvipsnames]{xcolor}

\usepackage[utf8]{inputenc}

\usepackage{textcomp}

\usepackage{tikz}
\usetikzlibrary{positioning}
\usetikzlibrary{decorations.pathreplacing}
\newcommand\myTab[1]{\begin{tabular}{c} #1 \end{tabular}}

\usepackage{algorithm}
\usepackage{algpseudocode}

\usepackage{letltxmacro}
\newlength{\commentindent}
\renewcommand{\Comment}[2][\commentindent]{%
  \leavevmode\hfill\makebox[#1][l]{$\triangleright$~#2}}

\usepackage{amssymb}
\usepackage{amsmath}
\usepackage{amsfonts}
\usepackage{mathtools}
\usepackage{mathrsfs}

\usepackage{enumerate}
\usepackage{enumitem}

\usepackage{fontawesome}

\usepackage{pgfplots}

\newcommand\ie{\textit{i.e.}, }
\newcommand\eg{\textit{e.g.}, }
\newcommand\Rtildestepj{$\widetilde{\mathbf{R}}_{[j[}$}
\newcommand\Rtildej{$\widetilde{\mathbf{R}}_j$}
\newcommand\Rtildejnext{$\widetilde{\mathbf{R}}_{j+1}$}
\newcommand\symbstatestepj{$([\mathbf{s}_{[j[}]_k, \mathbf{u}_{j,k})$}
\newcommand\symbstatej{$([\mathbf{s}_j]_k, \mathbf{u}_{j,k})$}

\newcommand\symbsj{$[\mathbf{s}_j]_k$}
\newcommand\symbsstepj{$[\mathbf{s}_{[j[}]_k$}
\newcommand\symbsjnext{$[\mathbf{s}_{j+1}]_k$}

\newtheorem{definition}{Definition}
\newtheorem{theorem}{Theorem}
\newtheorem{example}{Example}
\newtheorem{remark}{Remark}
\newtheorem{proof}{Proof}

\def\short{0}
\def\intermediate{1}

\begin{document}

\title{Safety Verification of Neural Network Controlled Systems}

\author{Arthur Clavi\`{e}re$^1$,
Eric Asselin$^1$,
Christophe Garion$^2$ and
Claire Pagetti$^3$
}

\date{$^1$\hspace*{1pt}Collins Aerospace, France \hspace*{5pt} $^2$\hspace*{1pt}ISAE-SUPAERO, France \hspace*{5pt} $^3$\hspace*{1pt}ONERA, France}

\maketitle              

\begin{abstract}
In this paper, we propose a system-level approach for verifying the safety of neural network controlled systems, combining a continuous-time physical system with a discrete-time neural network based controller. We assume a generic model for the controller that can capture both simple and complex behaviours involving neural networks. Based on this model, we perform a reachability analysis that soundly approximates the reachable states of the overall system, allowing to achieve a formal proof of safety. To this end, we leverage both validated simulation to approximate the behaviour of the physical system and abstract interpretation to approximate the behaviour of the controller. We evaluate the applicability of our approach using a real-world use case. Moreover, we show that our approach can provide valuable information when the system cannot be proved totally safe.
\end{abstract}

\section{Introduction}

Recently, feedforward deep neural networks have been successfully used for controlling physical systems, such as self-driving cars \cite{bojarski_2016,chen_2015,pan_2018} and unmanned aerial vehicles \cite{julian_2018}. The combination of a physical system with a neural network based controller is sometimes known as a \emph{neural network controlled system}. If such a system is considered as \emph{safety-critical}, meaning that a failure of the system could have serious consequences, then a particular effort needs to be made to demonstrate its safety. More precisely, one has to show evidence that the system fulfills a set of safety requirements, such as, in aeronautics, \textit{``A catastrophic failure shall occur with a probability less than $10^{-9}$ per hour of flight"}. 

Usually, to achieve this objective, the system has to be developped in accordance with stringent standards \eg ED-79A/ARP-4754A \cite{ARP-4754} in aeronautics. Such standards require several analyses to be performed, including safety assessment with fault trees. Moreover, together with these analyses, the system requirements have to be refined at the \emph{item level}, with the aim of achieving a \emph{correct}, \emph{comprehensive} specification for each item composing the system. Then, the development of each item must be performed in compliance with dedicated standards. For example, in aeronautics, the ED-12C/DO-178C \cite{DO-178} standard prescribes several verification activities to prove that a software item behaves \emph{exactly} as expected.

However, this classical approach is not applicable to neural network controlled systems. The reason for this is two fold. First, one cannot refine the system requirements at the neural network level. Indeed, most of the time, one cannot achieve a correct, comprehensive specification for the expected behaviour of a neural network. Generally, the expected behaviour of a network consists of a set of example data, which is a \emph{pointwise, non-comprehensive} specification. Secondly, existing standards such as ED-12C/DO-178C are not applicable to the development of neural network items. In particular, provided a comprehensive specification for the expected behaviour of a neural network can be defined, the learning process does not guarantee the correcteness of the resulting network. As a consequence, verifying that a network behaves exactly as expected may be infeasible, precisely because it does not.

To tackle these issues, we propose an alternative approach for demonstrating the safety of a neural network controlled system. This alternative approach aims at providing evidence that the overall system is safe, without performing item-level refinements and analyses. To this end, we leverage a model of the overall system, which accurately represents the items and their interactions. Then, a reachability analysis is performed on this model, with the aim of demonstrating that no reachable state can lead to a failure of the system.

The contributions of this paper are: (1) the definition of a \emph{realistic model} that can capture complex, real-world neural network controlled systems, involving one or more ReLU networks trained with supervised learning together with a pre-processing and a post-processing, (2) a reachability-based approach that allows to \emph{formally verify} the absence of errors leading to a failure of the system, and (3) an evaluation of the \emph{applicability} of our approach using a real-world use case.

The paper is organized as follows. Section \ref{sec:use_case} introduces the ACAS Xu use case, a real-world neural network controlled system that illustrates the applicability of our approach. Section \ref{sec:model} describes our model of a neural network controlled system and section \ref{sec:problem} defines the safety verification problem that we address. Section \ref{sec:approach} details our reachability analysis for solving the verification problem and section \ref{sec:experiments} presents the experimental results on the ACAS Xu use case. 

\section{Related work}

\paragraph{Neural network level} In the past few years, some progress has been made towards a more comprehensive specification for the expected behaviour of a neural network. Indeed, several research works have identified \emph{local} expected behaviours contributing to the overall expected behaviour of the network. Typically, a local behaviour consists of a pre-condition about the input of the network together with a post-condition about its output. An example of such a property is \emph{adversarial robustness} (also called local robustness) which captures the capability of the network to react correctly to a slight perturbation of a \emph{given} input \cite{katz_2017,huang_2020}. In recent years, there has been significant interest in verifying neural networks against this type of property, which has been shown to be a NP-complete problem \cite{katz_2017}. Several dedicated \emph {formal methods} have been proposed, with the advantage of providing a \emph{sound} analysis, meaning that the network is said correct only if it is actually correct. Some of these specialized formal methods are based on Satisfiability Modulo Theory solving \cite{katz_2017,ehlers_2017}, with the advantage of providing a \emph{complete} analysis \ie the network is said incorrect only if it is actually incorrect. However, these methods are often expensive for large, real-world sized networks. In order to offer a more \emph{scalable} analysis, other dedicated formal methods have been proposed, relying on abstract intepretation to soundly \emph{approximate} the semantics of the network \cite{wang_2018,gehr_2018,singh_2018,singh_2019}. Yet, as they consist of an over-approximation, these methods do not provide a complete analysis.

Our work does not address the safety objectives at the neural network level, so we do not seek to identify new local properties or to improve the existing verification techniques. However, we aim at using abstract interpretation based techniques to analyze the behaviour of the overall system. Indeed, such methods scale well to large networks and they provide not only a yes-or-no answer to a verification problem but also an approximation of the network semantics, that is helpful when reasoning about the overall system.

\paragraph{System level} Verifying the safety requirements at the system level, which corresponds to our approach, has been the object of a lot of insightful research. Indeed, there has been significant interest in verifying the safety of \emph{hybrid systems}, exhibiting both continuous-time and discrete-time dynamics \eg a physical system combined with a discrete-time controller. Among the proposed methods, \emph{falsification} aims at finding trajectories that violate a given safety property \cite{bogomolov_2019,annpureddy_2011}. Yet, even though falsification can prove that the system is unsafe, it cannot prove that the system is safe. \emph{Reachability analysis} can provide such a proof of safety by constructing a \emph{sound} approximation of the reachable states of the system and demonstrating that no reachable state can lead to a failure \cite{chen_2013,althoff_2015,sandretto_2015}. However, the classical reachability methods are not directly applicable to neural network controlled systems, due to the hardness of characterizing the input-output mapping of a neural network. Very recently, in the same vein as this paper, some research works have addressed the problem of verifying the safety of neural network controlled systems \cite{ivanov_2019,dutta_2019,huang_2019,tran_2020}. These works all assume a physical system combined with a periodically-scheduled controller that is a \emph{single} neural network \ie the input of the network is the sampled state of the physical system and the output of the network is the actuation command. To ensure the safety of such a system, they propose dedicated methods, all relying on reachability analysis.  

However, these methods are not applicable to \emph{complex} neural network controlled systems such as the ACAS Xu. Indeed, the controller used in the ACAS Xu is more sophisticated than a single neural network and the methods cited above cannot handle such a controller. \cite{julian_2019} has proposed an \emph{ad hoc} reachability approach for verifying the safety of the ACAS Xu at the system level, but the proposed method is not totally sound as it does not evaluate the reachable states for all instants but only for a set of \emph{discrete} instants. Moreover, it computes the reachable states by exploring the entire state space even though not all states are reachable. We propose here a \emph{generic} approach for \emph{soundly} verifying the safety of complex systems like the ACAS Xu, by exploring only the reachable states. 

\section{Use case}\label{sec:use_case}

\begin{figure}[hbt]
\begin{minipage}[c]{0.45\linewidth}
\centering
\def\angleOfIntruder{35} 
\def\distance{2.8} 
\def\sep{0.4}
\begin{tikzpicture}[scale=1.1]
\draw[->,draw=red,line width=1.2pt] (0:0) -- (90:1.1) node[left,pos=0.8]{\color{red}{$v_{\text{own}}$}};
\draw[->] (90:0.8) arc (90:\angleOfIntruder:0.8);
\draw (62:1) node {$\theta$};
\draw[dotted] (0:0) -- (\angleOfIntruder:\distance);
\node[inner sep=0] (n0) at (-55:\sep){};
\draw[dotted] (0:0) -- (n0);

\begin{scope}[shift={(\angleOfIntruder:\distance)}]
\draw[->,draw=red,line width=1.2pt] (0:0) -- (155:1.1) node[left,pos=1]{\color{red}{$v_{\text{int}}$}};
\draw[->] (90:0.8) arc (90:155:0.8);
\draw (120:1.1) node {$\psi$};
\draw[dotted] (0:0) -- (90:0.8);
\node[inner sep=0] (n1) at (-55:\sep){};
\draw[dotted] (0:0) -- (n1);
\end{scope}

\draw[<->] (n0) -- (n1) node[below,pos=0.5]{$\rho$};

\node[inner sep=0pt,rotate=45, scale=1.5] (plane1) at (0:0){\faPlane};
\node[inner sep=0pt,rotate=110, scale=1.5] (plane2) at (\angleOfIntruder:\distance){\faPlane};
\end{tikzpicture} 
\end{minipage} \hfill
\begin{minipage}[c]{0.45\linewidth}
\centering
\def\radE{0.4}
\def\radR{2}
\begin{tikzpicture}[scale=1]

\fill[Periwinkle!60]
  ([shift={(100:\radR)}]0,0) arc (100:110:\radR) 
  --
  (135:\radR)
  --
  (180:\radR)
  --
  ([shift={(225:\radR)}]0,0) arc (225:260:\radR)
   --
  (0:\radR*0.5)
  -- cycle;

\fill[white] plot [smooth cycle] coordinates { (110:\radR) (130:\radR*0.9) (180:\radR*0.8) (225:\radR) (220:\radR) (200:\radR*1.05) (180:\radR) (150:\radR) (135:\radR)};

\fill[white] plot [smooth cycle] coordinates { (100:\radR) (100:\radR*0.8) (135:\radR*0.4) (180:\radR*0.3) (225:\radR*0.4) (260:\radR) (265:\radR*1) (0:\radR) (100:\radR*1)};

\draw[draw=none, fill = BrickRed!50] (0,0) circle (\radE);
\draw (0:0) -- (45:\radE) node[right]{\tiny{$500$ ft}};
\node (n0) at (75:\radE + 0.2){\small{$\mathbf{E}$}};

\draw[dashed] (0,0) circle (\radR);
\node (n1) at (280:\radR*0.9){\small{$\mathcal{R}$}};

\fill[ForestGreen] plot [smooth cycle] coordinates { (100:\radR*1.05) (100:\radR*0.95) (111:\radR*0.95) (111:\radR*1.05)};
\node (n2) at (115:\radR*1.1){\small{$\mathbf{I}'$}};

\node[inner sep=0pt,rotate=45, scale=1.5] (plane1) at (0,0){\faPlane};

\end{tikzpicture} 
\end{minipage}
\caption{The 2D geometry of the encounter between the two aircraft (left) and the (illustrative) reachable trajectories of intruder relative to ownship from a subset $\mathbf{I}'$ of the possible initial states, with $\mathbf{E}$ representing a collision cylinder around the ownship and $\mathcal{R}$ delimiting the range of the ownship sensors (right).}
\label{fig:acasxu_use_case}
\end{figure}
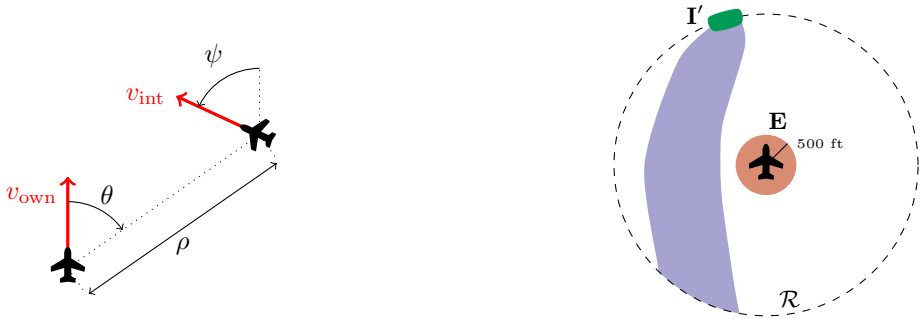

The safe integration of Unmanned Aerial Vehicles (UAVs) into the air traffic requires them to have collision avoidance capabilities. For this purpose, the standardization group RTCA SC 147 \cite{acas_standard} has recently developped a dedicated controller, namely the Airborne Collision Avoidance System for Unmanned Aircraft (ACAS Xu). The role of ACAS Xu is to avoid any collision between the \emph{ownship}, equipped with the controller, and an encountered aircraft called the \emph{intruder}, equipped or not with the controller. To this end, the ACAS Xu periodically provides the ownship with a \emph{horizontal maneuver advisory}, being either clear-of-conflict (COC), weak left turn (WL), weak right turn (WR), strong left turn (SL) or strong right turn (SR). The optimal advisory is extracted from a set of lookup tables, depending on the previous advisory and six variables describing the encounter between the two aircraft, defined in Fig. \ref{fig:acasxu_use_case}: (1) the distance $\rho$ from ownship to intruder, (2) the angle $\theta$ to intruder relative to ownship heading direction, (3) the heading angle $\psi$ of intruder relative to ownship heading direction, (4) the velocity $v_{\text{own}}$ of ownship, (5) the velocity $v_{\text{int}}$ of intruder and (6) the time $t_{\text{sep}}$ until loss of vertical separation. These six variables are computed from the input signals from the transponder and the sensors of the ownship \eg air-to-air radar, electro-optics/infrared sensors, cameras. The main weakness of the ACAS Xu controller is the associated storage requirements, over 2GB, which is too large for legacy avionics \cite{julian_2018}.

Recently, an alternative design for the ACAS Xu controller has been proposed, with dramatically reduced memory footprint (about 3 MB). It consists of a collection of $45$ neural networks approximating the lookup tables. Each single network approximates a table corresponding to a fixed previous advisory and a given interval for $t_{\text{sep}}$. As the possible values for $t_{\text{sep}}$ have been divided into $9$ intervals and $5$ possible advisories exist, the resulting controller uses $45$ networks. In addition to improving storage efficiency, this novel design also offers reduced runtime together with better performances, alerting the ownship earlier \cite{julian_2018}. However, due to the complexity of the neural networks composing the controller, we lack a proof that no collision can happen, whatever the initial state of the two aircraft (see Fig. \ref{fig:acasxu_use_case}). 

This use case will serve as an illustration of our approach in the rest of this paper. Our goal is to show evidence that the controller is effectively safe \ie it does prevent near mid-air collision.

\section{System model}\label{sec:model}

\if\intermediate0
This section details our model of a neural network controlled system. It also illustrates how the ACAS Xu can be represented by such a model.
\fi

\subsection{Closed-loop system}\label{sec:closed_loop}

We assume a \emph{closed-loop system} $\mathcal{C}$ that is the combination of a plant $\mathcal{P}$ and a neural network based controller $\mathcal{N}$. The plant $\mathcal{P}$ is a \emph{continuous-time} system while the controller $\mathcal{N}$ is a \emph{discrete-time} system, executed periodically with period $T$. They interact by means of a signal sampler and a zero-order-hold. More precisely:

\begin{itemize}[noitemsep,topsep=2pt]
\item[-] The state of the plant $\mathcal{P}$ at instant $t \in \mathbb{R}$ is the real-numbered vector $\mathbf{s}(t) \in \mathbb{R}^l$. The evolution of $\mathbf{s}(t)$ is continuous with $t$ and it depends, inter alia, on the actuation command from the controller, denoted by $\mathbf{u}(t) \in \mathbb{R}^d$.
\item[-] The $j^{\text{th}}$ execution of the controller (or control step) occurs in the time interval $[jT,(j+1)T[$. It takes as input the sampled state $\mathbf{s}_j = \mathbf{s}(jT)$ and it yields the command $\mathbf{u}_{j+1}$ to be applied for next period \ie $\mathbf{u}(t) = \mathbf{u}_{j+1} \ \forall t \in [(j+1)T, (j+2)T[$. This command $\mathbf{u}_{j+1}$ is taken from a \emph{finite} set $\mathbf{U} = \left\{\mathbf{u}^{(1)}, \ldots, \mathbf{u}^{(P)}\right\} \subset \mathbb{R}^d$, representing the possible actuation commands. It is worth noting that the controller is not assumed to execute instantaneously. Its execution time only has to be less than $T$, as for real systems.
\end{itemize}

\begin{figure}[hbt]
   \begin{center}
     \begin{tikzpicture}[pnt node/.style={circle,inner sep=0pt, minimum size=0pt, fill=red}]
     \small{
     \draw[very thin, dashed, rounded corners, fill=CornflowerBlue!10] (-4.5,-3.4) rectangle (4.5,-1.8);
     \node (t0) at (4.1,-3.2){$\mathcal{N}$};
      \draw[very thin,dashed, rounded corners, fill=gray!10] (-2.45,-0.9) rectangle (2.4,0.7);
      \node (t1) at (2.1,-0.7){$\mathcal{P}$};
     \node[rectangle, draw=black](n0)  at (0,0) {\myTab{ \textsc{Plant Dynamics} \\ $\mathbf{s}'(t) = f(t, \mathbf{s}(t), \mathbf{u}(t))$ }};
	\node[rectangle, draw=black, fill=yellow!5](n1) at (4.6, -1.0) { \myTab{\textsc{Sample} \\ \textsc{Hold}}};  
	\node[rectangle, draw=black](n2) at (3.1, -2.5) { \myTab{\textsc{Pre-} \\ \textsc{Processing}}};   
	\node[rectangle, draw=black](n3) at (0,-2.5) {\myTab{ \textsc{Neural Net.} \\ $\mathbf{y}_j = F_j(\mathbf{x}_j)$ }};  
	\node[rectangle, draw=black](n4) at (-3.1, -2.5) { \myTab{\textsc{Post-} \\ \textsc{Processing}}}; 
	\node[rectangle, draw=black, fill=yellow!5](n5) at (-4.6, -1.0) { \myTab{\textsc{Zero Order} \\ \textsc{Hold}}}; 
	
	\draw[line width=1.5pt, ->] (n0) -| node[above,pos=0.3]{$\mathbf{s}(t)$} (n1);
	\draw[line width=1.5pt, ->] (n1) |- node[right,pos=0.3]{$\mathbf{s}_j$} (n2);
	\draw[line width=1.5pt, ->] (n2) -- node[above]{$\mathbf{x}_j$} (n3);
	\draw[line width=1.5pt, ->] (n3) -- node[above]{$\mathbf{y}_j$} (n4);
	\draw[line width=1.5pt, ->] (n4) -|  node[left,pos=0.7]{$\mathbf{u}_{j+1}$}(n5);
	\draw[line width=1.5pt, ->] (n5) |- node[above,pos=0.7]{$\mathbf{u}(t)$}(n0);
	
	\node[pnt node](p0) at (5.55,-1.0) {};
	\node[pnt node](p1) at (-5.95,-1.0) {};
    \draw[line width=1.5pt, ->] (p0) edge  (n1) node[right]{\textsc{Clk}};
    \draw[line width=1.5pt, ->] (p1) edge  (n5) node[left]{\textsc{Clk}};
    }
     \end{tikzpicture}
   \end{center}
 \caption{Block diagram of a neural network closed-loop system $\mathcal{C}=\left(\mathcal{P},\mathcal{N}\right)$.}\label{fig:block-diagram}
\end{figure}
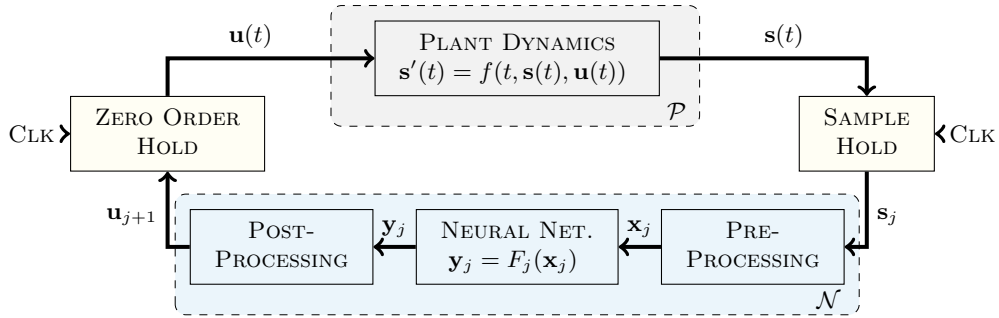

Overall, the state of the closed-loop system $\mathcal{C}$ is the $2$-tuple $\phi(t)=\left(\mathbf{s}(t), \mathbf{u}(t)\right)$ and we denote by $\phi_0 = \left(\mathbf{s}_0,\mathbf{u}_0\right) \in \mathbf{I}$ the initial state of $\mathcal{C}$, wherein $\mathbf{I} \subseteq \mathbb{R}^l \times \mathbf{U}$ is the set of the possible initial states. Moreover, we consider a set of \emph{erroneous} states $\mathbf{E} \subset \mathbb{R}^l \times\mathbf{U}$ such that a state $\phi(t) \in \mathbf{E}$ causes a potentially catastrophic failure of $\mathcal{C}$. It is thus expected that $\mathcal{C}$ does not reach a state in $\mathbf{E}$. We also assume that $\mathcal{C}$ terminates when its state $\phi(t)$ belongs to a set $\mathbf{T} \subset \mathbb{R}^l \times \mathbf{U}$, with $\mathbf{T} \cap \mathbf{E} = \emptyset$ to ensure a safe behaviour. Here, $\mathbf{T}$ can be seen as a set of \emph{target} states, corresponding to $\mathcal{C}$ having successfully achieved its mission. It is thus expected that $\mathcal{C}$ terminates in a \emph{finite} amount of time, whatever the initial state. We denote by $\tau \in \mathbb{R}$ the expected (or estimated) upper bound on this amount of time, independently of the initial state. Additionally, we set by definition $\phi(t) = \bot$ after the termination of the closed-loop system $\mathcal{C}$ \ie if $t_\text{end} \leq \tau$ satisfies $\forall t < t_{\text{end}}, \ \phi(t) \notin \mathbf{T}$ and $\phi(t_{\text{end}}) \in \mathbf{T}$ then $\phi(t) = \bot \ \forall t \in \ ]t_{\text{end}},\tau]$. In other words, the bottom element symbolically represents the ``terminated" state of $\mathcal{C}$.

Finally, as the combination of a deterministic plant $\mathcal{P}$ and a deterministic controller $\mathcal{N}$ (see sections \ref{sec:plant} and \ref{sec:controller}), the closed-loop system $\mathcal{C}$ has a deterministic behaviour. More precisely, for a given initial state $\phi_0 \in \mathbf{I}$, there exists a unique function $\phi_{\phi_0}: [0,\tau] \rightarrow \mathbb{R}^l \times \mathbf{U} \cup \{\bot\}$ such that $\phi_{\phi_0}(t)$ is the state of $\mathcal{C}$ at instant $t \leq \tau$. This hypothesis is important for properly defining the verification problem that we address, as well as demonstrating the soundness of our procedure.

\if\intermediate0
A more detailed desciption of both the plant $\mathcal{P}$ and the controller $\mathcal{N}$ is given in sections \ref{sec:plant} and \ref{sec:controller}.
\fi

\begin{example}
In the case of the ACAS Xu controller, we consider the plant $\mathcal{P}$ that is composed of both the ownship and the intruder. For simplicity, we assume that the two aircraft are at the same altitude, meaning that $t_{\text{sep}}$ equals $0$. Consequently, we define the state of $\mathcal{P}$ at instant $t$ as the real-numbered vector $\mathbf{s}(t) = \left( x(t) \ y(t) \ \psi(t) \ v_{\text{own}}(t) \ v_{\text{int}}(t)\right)^T$ where $x(t),y(t)$ are the 2D cartesian coordinates of intruder relative to ownship, $\psi(t)$ is the heading angle of intruder relative to ownship heading direction (measured counter clockwise), $v_{\text{own}}(t)$ and $v_{\text{int}}(t)$ denote the velocities of ownship and intruder respectively (see Fig. \ref{fig:acasxu_dynamics}). The neural network based controller $\mathcal{N}$ has a period $T=1s$.  It outputs the actuation command $u(t) \in \mathbb{R}$ that is the turn rate of ownship, measured counter clockwise. This command is taken from the set $\mathbf{U}=\{0 \ \text{deg/s},1.5 \ \text{deg/s},-1.5 \ \text{deg/s},3 \ \text{deg/s},-3 \ \text{deg/s}\}$, of which values represent COC, WL, WR, SL and SR respectively. Overall, an initial state $\phi_0 = \left(\mathbf{s}_0,u_0\right)$ of the closed-loop $\mathcal{C}$ corresponds to the intruder being detected by ownship for the first time. Therefore, the initial position $(x_0, y_0)$ of intruder lies along a circle $\mathcal{R}$ centered on ownship and with a radius $r$ equal to the range of the ownship sensors (see Fig. \ref{fig:acasxu_use_case}). Here we consider that $r=8000 \ \text{ft}$, which is a reasonable hypothesis. Furthermore, the initial angle $\psi_0$ is such that the intruder penetrates the circle $\mathcal{R}$ \textit{i.e.} $\psi_0$ belongs to a cone delimited by the tangent to $\mathcal{R}$ at the point $(x_0,y_0)$. The initial actuation command $u_0$ is $0.0 \ \text{deg/s}$, corresponding to a Clear-of-Conflict, and we assume for simplicity that $v_{\text{own},0}=700.0 \ \text{ft/s}$ and $v_{\text{int},0}=600.0 \ \text{ft/s}$. The set of the possible initial states $\mathbf{I}$ is thus defined by the set of the possible tuples $(x_0,y_0,\psi_0)$. Additionally, we consider a set $\mathbf{E}$ of erroneous states representing a collision between the two aircraft. Such a collision happens when the intruder enters the collision circle around ownship, with a radius of $500 \ \text{ft}$ \cite{manfredi_2016}, hence $\mathbf{E} = \{ \phi(t)=(\mathbf{s}(t),u(t)) \in \mathbb{R}^l \times \mathbf{U} \ | \ \sqrt{\ x(t)^2 + y(t)^2} < 500.0 \ \text{ft}\}$. Finally, the closed-loop system terminates when the intruder leaves the circle $\mathcal{R}$ \textit{i.e.} the ownship does not see it anymore: $\mathbf{T} = \{ \phi(t)=(\mathbf{s}(t),u(t)) \in \mathbb{R}^l \times \mathbf{U} \ | \ \sqrt{\ x(t)^2 + y(t)^2} > r \}$. As the two aircraft have different velocities, it is expected that $\mathcal{C}$ terminates in a finite amount of time. We take $\tau = 20\text{s}$ as the upper bound on this amount of time, which is relevant given the values of $v_{\text{own}}$, $v_{\text{int}}$ and $r$.
\end{example}

\subsection{Plant dynamics}\label{sec:plant}

The dynamics of the plant $\mathcal{P}$ \ie the temporal evolution of its state $\mathbf{s}(t)$, is modelled by an \emph{ordinary differential equation}. 

\begin{definition} An ordinary differential equation (ODE) is a relation between a function $\mathbf{z}: \mathbb{R} \rightarrow \mathbb{R}^l, \ t \mapsto \mathbf{z}(t)$ and its derivative $\mathbf{z}' = \frac{d\mathbf{z}}{dt}$ of the form $\mathbf{z}'(t) = f(t,\mathbf{z}(t))$ wherein $f: \mathbb{R} \times \mathbb{R}^l \rightarrow \mathbb{R}^l$. 
\end{definition}

To take account of the command signal $\mathbf{u}(t)$, the dynamics of $\mathcal{P}$ is of the form $\mathbf{s}'(t) = f(t,\mathbf{s}(t),\mathbf{u}(t))$ wherein $f: \mathbb{R} \times \mathbb{R}^l \times \mathbb{R}^d \rightarrow \mathbb{R}$ is assumed to be continuous in $t$ and $\mathbf{u}$ and uniformly Lipschitz continuous in $\mathbf{s}$  \ie its slope \textit{w.r.t.} $\mathbf{s}$ is uniformly bounded on $\mathbb{R} \times \mathbb{R}^l \times \mathbb{R}^d$. Indeed, under these hypotheses and when $\mathbf{u}(t)$ is a piecewise constant function (as in the case of $\mathcal{C}$), then $\mathcal{P}$ has a deterministic behaviour. More precisely, let us consider a time interval $[0,qT]$ with $q \in \mathbb{N}$ and a given command signal $\mathbf{u}(t)$, constant on $[jT,(j+1)T[$ for $j < q$. There exists a unique function $\mathbf{s}^*$ defined on $[0,qT]$, continuous on $[0,qT]$, such that it verifies the ODE on each open interval $]jT,(j+1)T[$ for $j < q$, and the initial condition $\mathbf{s}(0) = \mathbf{s}_0$. 

\if\short0
\begin{proof} The function $\mathbf{s}^*$ can be constructed iteratively. The initial condition imposes $\mathbf{s}^*(0) = \mathbf{s}_0$. Then, for $j \in [\![0,q-1]\!]$, the Picard-Lindelöf theorem ensures the existence and uniqueness of a function $\mathbf{s}^*_j$ satisfying $\mathbf{s}'(t) = f(t,\mathbf{s}(t), \mathbf{u}(jT))$ and $\mathbf{s}(jT) = s^*(jT)$. In order for $\mathbf{s}^*$ to be continuous at instants $t=jT$ and $t=(j+1)T$ and to satisfy the ODE on $]jT,(j+1)T[$, it must be such that $\mathbf{s}^*(t) = \mathbf{s}^*_j(t) \ \forall \ t \in \ ]jT,(j+1)T]$. Hence the existence and uniqueness of $\mathbf{s}^*$. $\square$
\end{proof}

\begin{remark}
The function $\mathbf{s}^*$ is not derivable at instants $t=T,2T,\ldots,(q-1)T$. This is not quite realistic from a physical point of view as it means that the plant $\mathcal{P}$ reacts instantaneously to a new actuation command. However, this is a common hypothesis when modelling such a system. 
\end{remark}
\fi

\begin{figure}[hbt]
\begin{minipage}[c]{0.40\linewidth}
\raggedleft
\def\xown{0}
\def\yown{0}
\def\xint{2.3}
\def\yint{1.6}
\def\sep{0.4}
\begin{tikzpicture}[scale=1.1]
\draw[dotted] (\xown,\yown) -- (\xown,\yown - 0\sep);
\draw[dotted] (\xown,\yown) -- (\xint + \sep,\yown);
\begin{scope}[shift={(\xown,\yown)}]
\draw[->,draw=red,line width=1.2pt] (0:0) -- (90:1.1) node[left,pos=0.8]{\color{red}{$v_{\text{own}}$}};
\end{scope}

\draw[dotted] (\xint,\yint) -- (\xint,\yown - \sep);
\draw[dotted] (\xint,\yint) -- (\xint + \sep,\yint);
\draw[dotted] (\xint,\yint) -- (\xint,\yint+0.8);
\begin{scope}[shift={(\xint,\yint)}]
\draw[->,draw=red,line width=1.2pt] (0:0) -- (155:1.1) node[left,pos=1]{\color{red}{$v_{\text{int}}$}};
\draw[->] (90:0.8) arc (90:155:0.8);
\draw (120:1.1) node {$\psi$};
\end{scope}

\draw[->] (\xown,\yown - \sep) -- (\xint,\yown - \sep) node[below,pos=0.5]{$x$};
\draw[->] (\xint + \sep,\yown) -- (\xint + \sep,\yint) node[right,pos=0.5]{$y$};

\node[inner sep=0pt,rotate=45, scale=1.5] (plane1) at (\xown,\yown){\faPlane};
\node[inner sep=0pt,rotate=110, scale=1.5] (plane2) at (\xint,\yint){\faPlane};
\end{tikzpicture} 
\end{minipage} \hfill
\begin{minipage}[c]{0.55\linewidth}
\raggedright
\begin{equation}\label{eq:acasxu_ode}
\left\{
\begin{split}
x'(t) &=  - v_{\text{int}}(t) \cdot \text{sin}(\psi(t))\\
y'(t) &= v_{\text{int}}(t) \cdot \text{cos}(\psi(t)) - v_{\text{own}}(t)\\
\psi'(t) &= -u(t)\\
v_{\text{own}}'(t) &= 0\\
v_{\text{int}}'(t) &=0\\ 
\end{split}
\right.
\end{equation}
\end{minipage}
\caption{The 2D kinematic model of the plant $\mathcal{P}$ of the ACAS Xu use case, composed of both the ownship and the intruder.}
\label{fig:acasxu_dynamics}
\end{figure}
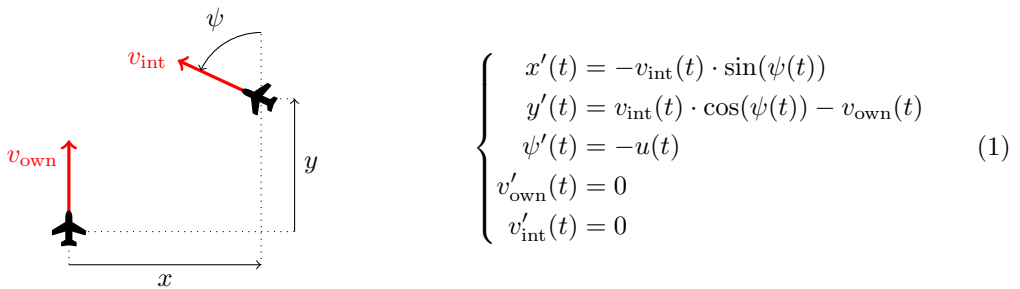

\begin{example}
For the ACAS Xu, the temporal evolution of $\mathbf{s}(t)$ is modelled by the ODE $\mathbf{s}'(t) = f(t,\mathbf{s}(t),\mathbf{u}(t))$ given in equation (\ref{eq:acasxu_ode}). This ODE is based on a 2D \emph{non-linear} kinematic model where the intruder is assumed to keep constant heading and velocity: the evolution of $\psi(t)$ depends only on the evolution of the ownship heading. This corresponds to a degraded mode where the intruder does not perform any collision avoidance maneuver and continues its uniform rectilinear displacement. For simplicity, the velocity of ownship is also considered constant. It is worth noting that $f$ is continuous in $t$ and $u$, as well as uniformly Lipschitz continuous in $\mathbf{s}$. Indeed, its derivative \textit{w.r.t.} $\mathbf{s}$ is bounded on $\mathbb{R} \times \mathbb{R}^l \times \mathbb{R}^d$ since both $v_{\text{own}}(t)$ and $v_{\text{int}}(t)$ are constants.
\end{example}

\subsection{Neural network based controller}\label{sec:controller}

The neural network based controller $\mathcal{N}$ involves a collection of \emph{ReLU neural networks} $\mathbf{N} = \left\{N^{(1)},\ldots,N^{(D)}\right\}$, of which only one is executed at each control step. The network $N_j \in \mathbf{N}$ to be executed at step $j$ is selected based on the command $\mathbf{u}_j$ produced at previous step \ie $N_j=\lambda(\mathbf{u}_j)$ wherein $\lambda: \mathbf{U} \rightarrow \mathbf{N}$ maps every command in $\mathbf{U}$ to a network in $\mathbf{N}$. It is worth noting that all the neural networks in $\mathbf{N}$ are assumed to have been trained already, meaning that they remain unchanged for the run-time of the controller. 

\begin{definition}
A \emph{ReLU feedforward deep neural network} is a tuple $N=(L, \{k_l\}_{1\leq l \leq L}, \mathbf{W},\mathbf{B})$. It consists of a \emph{directed acyclic weighted graph} where the nodes are arranged in $L$ layers, comprising $k_1,\ldots, k_L$ nodes respectively. The first layer is called the \emph{input layer}, the last layer is called the \emph{output layer}, and the layers in between are called the \emph{hidden layers}. Except the input layer, each layer has its nodes connected to the nodes in the preceding layer. More precisely, let $n_{l,i}$ be the $i^{th}$ node in the $l^{th}$ layer. If $l >1$, there exists an edge from $n_{l-1,j}$ to $n_{l,i}$ for each $i \in [\![1,k_l]\!]$ and $j \in [\![1,k_{l-1}]\!]$. Moreover, the edge from $n_{l-1,j}$ to $n_{l,i}$ is assigned a \emph{weight} $w_{l,i}^j \in \mathbf{W}$ and each non-input node $n_{l,i}$ is assigned a \emph{bias} $b_{l,i} \in \mathbf{B}$.

This graph actually corresponds to a function $F:\mathbb{R}^{k_1} \rightarrow \mathbb{R}^{k_L}$. Indeed, each node $n_{l,i}$ represents a function $F_{l,i}$ of which definition depends on the layer $l$. For the nodes in the input layer, this function is the identity function \ie $F_{1,i}\triangleq {\textnormal{id}}_{\mathbb{R}}, \ \forall i \in [\![1,k_1]\!]$. For the nodes in the hidden layer $l$, with $1 < l < L$, the associated function maps a vector in $\mathbb{R}^{k_{l-1}}$ to an element in $\mathbb{R}$. It is the composition of a \emph{non-linear ReLU unit} $\sigma: x \mapsto \textnormal{max}(0,x)$ and an \emph{affine transformation} \ie $F_{l,i}: \mathbf{z} \mapsto
\sigma \left( \sum_{j=1}^{k_{l-1}} w_{l,i}^j \cdot \mathbf{z}_j + b_{l,i}\right), \ \forall i \in [\![1,k_l]\!]$. Finally, the function represented by the nodes in the output layer is an affine transformation of a vector in $\mathbb{R}^{k_{L-1}}$ \ie $F_{L,i}: \mathbf{z} \mapsto
\sum_{j=1}^{k_{L-1}} w_{L,i}^j \cdot \mathbf{z}_j + b_{L,i}, \ \forall i \in [\![1,k_L]\!]$. Overall, the function computed by the $l^{th}$ layer of the network is the vector function $F_l: \mathbf{z} \mapsto \left( F_{l,1}(\mathbf{z}) \ \ldots \ F_{l,k_l}(\mathbf{z})\right)^T$ and the function $F$ computed by the network is the composition function $F \triangleq F_L \circ \ldots \circ F_1$. In particular, $F$ is a deterministic function.

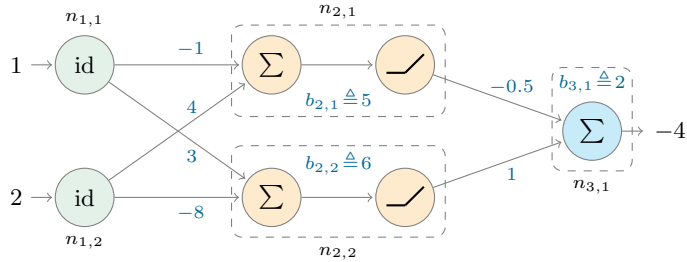
\begin{figure}[hbt]
\centering
\def\layersep{70pt}
\def\hiddensep{50}
\def\neuronsep{50pt}
\def\inoutsep{20pt}
\def\posweight{0.6}
\def\margin{15pt}
\def\relusize{7pt}
\def\reluoffset{3pt}
\begin{tikzpicture}[scale=1,draw=black!50, node distance=\layersep]
\small{
    \tikzstyle{every pin edge}=[<-,shorten <=1pt]
    \tikzstyle{neuron}=[circle,draw,fill=black!25,minimum size=22pt,inner sep=0pt]
    \tikzstyle{input neuron}=[neuron, fill=ForestGreen!10];
    \tikzstyle{output neuron}=[neuron, fill=ProcessBlue!20];
    \tikzstyle{hidden neuron}=[neuron, fill=YellowOrange!20];
    \tikzstyle{hidden bias}=[circle,draw,dashed,minimum size=22pt,inner sep=0pt,fill=ProcessBlue!20];
    \tikzstyle{output bias}=[circle,draw,dashed,minimum size=22pt,inner sep=0pt,fill=red!15];
    \tikzstyle{annot} = [text width=4em, text centered]
    
    \draw[thin, dashed, rounded corners] (\layersep-\margin, -1*\neuronsep-1.3*\margin) rectangle (\layersep+\hiddensep+\margin,-1*\neuronsep+\margin);  
    \draw[thin, dashed, rounded corners] (\layersep-\margin, -2*\neuronsep-\margin) rectangle (\layersep+\hiddensep+\margin,-2*\neuronsep+1.3*\margin); 
     \draw[thin, dashed, rounded corners] (2*\layersep+\hiddensep-\margin, -3/2*\neuronsep-\margin) rectangle (2*\layersep+\hiddensep+\margin,-3/2*\neuronsep+1.6*\margin);

    \foreach \name / \y in {1,...,2}
        \node[input neuron] (I-\name) at (0,-\y*\neuronsep) {id};
                
   \node[above = 0pt of I-1] {\scriptsize{$n_{1,1}$}};
    \node[below = 0pt of I-2] {\scriptsize{$n_{1,2}$}};
        
    \draw[->,shorten >=1pt] (-\inoutsep,-1*\neuronsep) -- (I-1) node[left, pos=0] {$1$};
    \draw[->,shorten >=1pt] (-\inoutsep,-2*\neuronsep) -- (I-2) node[left, pos=0] {$2$};

    \foreach \name / \y in {1,...,2}
        \path[yshift=0pt]
            node[hidden neuron] (Hin-\name) at (\layersep,-\y*\neuronsep) {$\sum$};
            
    \foreach \name / \y in {1,...,2}
        \path[yshift=0pt]
            node[hidden neuron] (Hout-\name) at (\layersep + \hiddensep,-\y*\neuronsep) {};
            
     \draw[->, shorten >=1pt] (Hin-1) -- (Hout-1) node[pos=0.5] (h-1) {};
	\node[above = 11pt of h-1] {\scriptsize{$n_{2,1}$}};
	\node[below = 2pt of h-1] {\scriptsize{\textcolor{MidnightBlue}{$b_{2,1} \! \triangleq \! 5$}}};
	
	\draw[->, shorten >=1pt] (Hin-2) -- (Hout-2) node[pos=0.5] (h-2) {};
	\node[below = 12pt of h-2] {\scriptsize{$n_{2,2}$}};
	\node[above = 2pt of h-2] {\scriptsize{\textcolor{MidnightBlue}{$b_{2,2} \! \triangleq \! 6$}}};

    \path[yshift=0pt]
    	node[output neuron] (O) at (2*\layersep + \hiddensep,-3/2*\neuronsep) {$\sum$};
    
    \draw[->,shorten >=1pt] (O) -- (2*\layersep+\hiddensep+\inoutsep,-3/2*\neuronsep) node[right, pos=1] {$- 4$};
    
   \node[below= 4pt of O] {\scriptsize{$n_{3,1}$}};
	\node[above = 0pt of O] {\scriptsize{\textcolor{MidnightBlue}{$b_{3,1} \! \triangleq \! 2$}}};
    	
    \draw[->, shorten >=1pt] (I-1) -- (Hin-1) node[above, pos=\posweight] {\scriptsize{\textcolor{MidnightBlue}{$-1$}}};
    \draw[->, shorten >=1pt] (I-1) -- (Hin-2) node[below,pos=\posweight] {\scriptsize{\textcolor{MidnightBlue}{$3$}}};
    \draw[->, shorten >=1pt] (I-2) -- (Hin-1) node[above,pos=\posweight] {\scriptsize{\textcolor{MidnightBlue}{$4$}}};
    \draw[->, shorten >=1pt] (I-2) -- (Hin-2) node[below,pos=\posweight] {\scriptsize{\textcolor{MidnightBlue}{$-8$}}};
	\draw[->, shorten >=1pt] (Hout-1) -- (O) node[above,pos=\posweight] {\scriptsize{\textcolor{MidnightBlue}{$-0.5$}}};
	\draw[->, shorten >=1pt] (Hout-2) -- (O) node[below,pos=\posweight] {\scriptsize{\textcolor{MidnightBlue}{$1$}}};
	
	\draw[thick, draw=black] (\layersep+\hiddensep-\relusize,-1*\neuronsep-\reluoffset) -- (\layersep+\hiddensep,-1*\neuronsep-\reluoffset);
	\draw[thick, draw=black] (\layersep+\hiddensep,-1*\neuronsep-\reluoffset) -- (\layersep+\hiddensep+\relusize,-1*\neuronsep-\reluoffset+\relusize);
	
	\draw[thick, draw=black] (\layersep+\hiddensep-\relusize,-2*\neuronsep-\reluoffset) -- (\layersep+\hiddensep,-2*\neuronsep-\reluoffset);
	\draw[thick, draw=black] (\layersep+\hiddensep,-2*\neuronsep-\reluoffset) -- (\layersep+\hiddensep+\relusize,-2*\neuronsep-\reluoffset+\relusize);
	
    
    
	
 }
\end{tikzpicture}
\caption{A (tiny) example ReLU network $N = \left(3,\{2,2,1\},\mathbf{W}, \mathbf{B}\right)$. }\label{fig:nn}
\end{figure}

In the example of Fig. \ref{fig:nn}, the input layer yields $F_1\left((1 \ 2)\right) = (1 \ 2)$, then the hidden layer yields $F_2\left((1 \ 2)\right) = \left(\sigma( - \! 1 \! \times \! 1 \! + \! 4 \! \times \! 2 \! + \! 5) \ \sigma( 3 \! \times \! 1 \! - \! 8 \! \times \! 2 \! + \! 6)\right) = (12 \ 0)$, and the output layer yields $F_3\left((12 \ 0)\right) = (- 0.5 \! \times \! 12 \! + \! 1 \! \times \! 0 \! + \! 2) = -4$.
\end{definition}

In addition to the neural networks, the controller $\mathcal{N}$ involves both a pre-processing and a post-processing stage. More precisely, the $j^{th}$ execution of the controller consists of: (i) a \emph{pre-processing} which calculates the input $\mathbf{x}_j \in \mathbb{R}^m$ of the network $N_j$ \ie $\mathbf{x}_j = \text{Pre}(\mathbf{s}_j)$ wherein $\text{Pre}:\mathbb{R}^l \rightarrow \mathbb{R}^m$ (\eg calculation of a distance from two positions, normalization) (ii) the \emph{neural network execution}, which yields the output vector $\mathbf{y}_j \in \mathbb{R}^p$ such that $\mathbf{y}_j = F_j(\mathbf{x}_j)$ where $F_j : \mathbb{R}^m \rightarrow \mathbb{R}^p$ is the function computed by the network $N_j$, and (iii) a \emph{post-processing} which determines the command $\mathbf{u}_{j+1}$ given the neural network output $\mathbf{y}_j$ \ie $\mathbf{u}_{j+1}=\text{Post}(\mathbf{y}_j)$ where $\text{Post}:\mathbb{R}^p \rightarrow \mathbf{U}$. Typically, each component $\left(\mathbf{y}_j\right)_i \in \mathbb{R}$ of the output $\mathbf{y}_j$ of the network could correspond to a command $\mathbf{u}^{(i)} \in \mathbf{U}$, and the post-processing be $\mathbf{u}_{j+1} = \mathbf{u}^{(k)} \ \text{\textit{s.t.}} \quad k = \underset{i}{\text{argmin}}\left(\left(\mathbf{y}_j\right)_i\right)$. 

Both the pre and post processing are assumed to be deterministic functions, so that the whole controller is also a deterministic function. The overall architecture of $\mathcal{N}$ is illustrated in Fig. \ref{fig:block-diagram}.

\begin{example}
To decide on the maneuver to perform, the ACAS Xu controller uses a collection of $5$ ReLU networks $\mathbf{N} = \left\{ N^{(1)}, \ldots, N^{(5)} \right\}$. These networks all have $6$ hidden layers of $50$ nodes each. They were each trained with supervised learning to approximate a table of the original ACAS Xu, corresponding to one of the $5$ possible previous advisories and $t_{\text{sep}}=0$ (the $40$ remaining networks are not considered as they correspond to $t_{\text{sep}} \neq 0$). Therefore, the function $\lambda$ selecting the network to be executed maps the $5$ possible advisories to the $5$ networks in $\mathbf{N}$. The pre-processing stage transforms the sampled state $\mathbf{s}_j$ into the input $\mathbf{x}_j$ of the network by replacing the cartesian coordinates $x,y$ into the cylindrical coordinates $\rho,\theta$ (defined in Fig. \ref{fig:acasxu_use_case}), and normalizes the resulting vector. The function $F_j:\mathbb{R}^5 \rightarrow \mathbb{R}^5$ computed by the neural network then outputs $5$ scores, each one corresponding to a possible maneuver. Finally, the post-processing consists of a argmin function: it chooses the maneuver with the minimal score. A model of the ACAS Xu controller is given in Fig. \ref{fig:acasxu_controller}.

\begin{figure}[hbt]
\centering
\def\offset{0.2}
\def\hsep{1}
\def\inhsep{0.3}
\def\vsep{0.7}
\def\bwidth{1.2}
\def\bheight{3.2}
\begin{tikzpicture}
\draw[fill=white] (0,0) rectangle (\bwidth,\bheight);
\node[rotate=90](n0) at (\bwidth/2,\bheight/2) {\myTab{ \textsc{Pre Processing}}};  
\draw[->,draw] (-\inhsep,\offset + 4*\vsep) -- (0,\offset + 4*\vsep) node[left,pos=0]{$x_j$};
\draw[->,draw] (-\inhsep,\offset + 3*\vsep) -- (0,\offset + 3*\vsep) node[left,pos=0]{$y_j$};
\draw[->,draw] (-\inhsep,\offset + 2*\vsep) -- (0,\offset + 2*\vsep) node[left,pos=0]{$\psi_j$};
\draw[->,draw] (-\inhsep,\offset + 1*\vsep) -- (0,\offset + 1*\vsep) node[left,pos=0]{$v_{\text{own},j}$};
\draw[->,draw] (-\inhsep,\offset + 0*\vsep) -- (0,\offset + 0*\vsep) node[left,pos=0]{$v_{\text{int},j}$};
\draw[fill=white] (\bwidth + \hsep,0) rectangle (2*\bwidth + \hsep,\bheight);
\node[rotate=90](n0) at (\bwidth + \hsep + \bwidth/2,\bheight/2) {\myTab{ \textsc{Neural Net. $N_j$}}};  
\draw[->,draw] (\bwidth,\offset + 4*\vsep) -- (\bwidth + \hsep,\offset + 4*\vsep) node[above,pos=0.5]{$\rho_j$};
\draw[->,draw] (\bwidth,\offset + 3*\vsep) -- (\bwidth + \hsep,\offset + 3*\vsep) node[above,pos=0.5]{$\theta_j$};
\draw[->,draw] (\bwidth,\offset + 2*\vsep) -- (\bwidth + \hsep,\offset + 2*\vsep) node[above,pos=0.5]{$\psi_j$};
\draw[->,draw] (\bwidth,\offset + 1*\vsep) -- (\bwidth + \hsep,\offset + 1*\vsep) node[above,pos=0.5]{$v_{\text{own},j}$};
\draw[->,draw] (\bwidth,\offset + 0*\vsep) -- (\bwidth + \hsep,\offset + 0*\vsep) node[above,pos=0.5]{$v_{\text{int},j}$};
\draw[fill=white] (2*\bwidth + 2*\hsep,0) rectangle (3*\bwidth + 2*\hsep,\bheight);
\node[rotate=90](n0) at (2*\bwidth + 2*\hsep + \bwidth/2,\bheight/2) {\myTab{ \textsc{Post Processing}}};  
\draw[->,draw] (2*\bwidth + \hsep,\offset + 4*\vsep) -- (2*\bwidth + 2*\hsep,\offset + 4*\vsep) node[above,pos=0.5]{\small{$\text{COC}_j$}};
\draw[->,draw] (2*\bwidth + \hsep,\offset + 3*\vsep) -- (2*\bwidth + 2*\hsep,\offset + 3*\vsep) node[above,pos=0.5]{\small{$\text{WL}_j$}};
\draw[->,draw] (2*\bwidth + \hsep,\offset + 2*\vsep) -- (2*\bwidth + 2*\hsep,\offset + 2*\vsep) node[above,pos=0.5]{\small{$\text{WR}_j$}};
\draw[->,draw] (2*\bwidth + \hsep,\offset + 1*\vsep) -- (2*\bwidth + 2*\hsep,\offset + 1*\vsep) node[above,pos=0.5]{\small{$\text{SL}_j$}};
\draw[->,draw] (2*\bwidth + \hsep,\offset + 0*\vsep) -- (2*\bwidth + 2*\hsep,\offset + 0*\vsep) node[above,pos=0.5]{\small{$\text{SR}_j$}};
\draw[->,draw] (3*\bwidth + 2*\hsep,\offset + 2*\vsep) -- (3*\bwidth + 2*\hsep + \inhsep,\offset + 2*\vsep) node[right,pos=1]{$u_{j+1}$};
\end{tikzpicture}
\caption{Model of the neural network based ACAS Xu controller.}
\label{fig:acasxu_controller}
\end{figure}
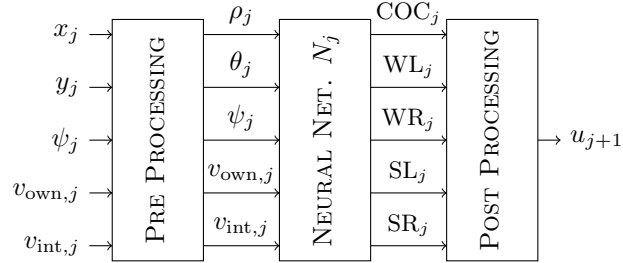
\end{example}

\section{Safety verification problem}\label{sec:problem}

In this section, we consider the closed-loop system $\mathcal{C}$ and its evolution over the time horizon $\tau$, the purpose being to prove that no unsafe state can be reached over $[0,\tau]$.

\subsection{Reachability definition}\label{sec:reachability}
Given the deterministic behaviour of the closed-loop system $\mathcal{C}$ (see section \ref{sec:closed_loop}), we define the reachable states of $\mathcal{C}$ as follows:

\begin{definition}\label{def:Rt}
The reachable states of the closed-loop system $\mathcal{C}$ at a given instant $t\leq\tau$ is the set $\mathbf{R}_{t}  = \{ \phi \in \mathbb{R}^l \times \mathbf{U} \cup \{\bot\}\ | \ \exists \phi_0 \in \mathbf{I}, \ \phi = \phi_{\phi_0}(t)\}$ (see section \ref{sec:closed_loop} for the definition of $\phi_{\phi_0}$). 
\end{definition}

\begin{definition}\label{def:Rtt}
The reachable states of the closed-loop system $\mathcal{C}$ for the time interval $[t_1,t_2] \subset [0,\tau]$ (resp. $[t_1,t_2[ \subset [0,\tau]$) is the set $\mathbf{R}_{[t_1,t_2]} = \{ \phi \in \mathbf{R}_t \ | \ t \in [t_1,t_2]\}$ (resp. $\mathbf{R}_{[t_1,t_2[} = \{ \phi \in \mathbf{R}_t \ | \ t \in [t_1,t_2[\}$).
\end{definition}

\subsection{Problem definition}

We want to \emph{decide} if, whatever the initial state $\phi_0$ in $\mathbf{I}$,  the closed-loop system $\mathcal{C}$ remains safe \textit{w.r.t} the set of erroneous states $\mathbf{E}$ over the time horizon $\tau$. In other words, we want to decide if the reachable states of $\mathcal{C}$ in $[0,\tau]$ remain outside $\mathbf{E}$. 

\begin{definition}
The \emph{safety verification problem} $\mathcal{V}$ consists in deciding if:
\begin{equation}\label{eq:problem}
\mathbf{R}_{[0,\tau]} \cap \mathbf{E} = \emptyset
\end{equation}
\end{definition}

\paragraph{}
Reasoning about the problem $\mathcal{V}$ is a difficult task. Indeed, whatever the nature of the controller (based on ReLU networks or not), the problem $\mathcal{V}$ is undecidable when the plant $\mathcal{P}$ has a non-linear dynamics \cite{alur_1995,hainry_2008} (\eg ACAS Xu). Furthermore, the neural networks add to the complexity of the verification problem. Indeed, due to the \emph{non-linear} ReLU units and the \emph{many dependencies} induced by the affine transformations, the function computed by a ReLU network is non-monotonic, non convex and highly non-linear. As a result, its behaviour is very difficult to analyze for a \emph{continuum} of inputs, which is the case in problem $\mathcal{V}$ as the initial set $\mathbf{I}$ is infinite. Actually, it has been shown that verifying pre/post-conditions on a ReLU network is a NP-hard problem \cite{katz_2017}. Finally, the controller we consider has a non-trivial logic, switching between the networks and involving pre and post-processing stages, which increases the dependencies from one control step to another. 

\paragraph{}
As the problem $\mathcal{V}$ is undecidable, we aim at constructing a \emph{sound} approximation of the reachable states of $\mathcal{C}$. More precisely, we aim at computing a \emph{bounded} set $\widetilde{\mathbf{R}}_{[0,\tau]}$ satisfying $\widetilde{\mathbf{R}}_{[0,\tau]} \supset \mathbf{R}_{[0,\tau]}$. Indeed, provided we are able to compute such a set and if it verifies $\widetilde{\mathbf{R}}_{[0,\tau]} \cap \mathbf{E} = \emptyset$, then (\ref{eq:problem}) is proved to hold. Consequently, we consider the problem $\widetilde{\mathcal{V}}$ defined as follows:

\begin{definition}
The \emph{safety verification problem} $\widetilde{\mathcal{V}}$ consists in finding a set $\widetilde{\mathbf{R}}_{[0,\tau]}$ satisfying $\widetilde{\mathbf{R}}_{[0,\tau]} \supset \mathbf{R}_{[0,\tau]}$ and $\widetilde{\mathbf{R}}_{[0,\tau]} \cap \mathbf{E} = \emptyset$.
\end{definition}

To have a chance to find a solution to problem $\widetilde{\mathcal{V}}$, the set $\widetilde{\mathbf{R}}_{[0,\tau]}$ must be as tight as possible. The next section presents our method for computing a \emph{tight} over-approximation $\widetilde{\mathbf{R}}_{[0,\tau]} \supset \mathbf{R}_{[0,\tau]}$.

\section{Reachability-based approach}\label{sec:approach}

\subsection{Symbolic state and symbolic set}

The set $\widetilde{\mathbf{R}}_{[0,\tau]}$ that we aim at constructing is \emph{infinite}. To allow reasoning about this type of set, we introduce the notions of \emph{symbolic state} and \emph{symbolic set}.

\begin{definition}
A \emph{symbolic state} is a $2$-tuple $([\mathbf{s}],\mathbf{u})$ wherein $[\mathbf{s}] \subset \mathbb{R}^l$ is a $l$-dimensional box \ie the cartesian product of $l$ intervals, and $\mathbf{u} \in \mathbf{U}$. It symbolically represents the set $\{ \phi(t)=(\mathbf{s}(t),\mathbf{u}(t)) \in \mathbb{R}^l \times \mathbf{U} \ | \ \mathbf{s}(t) \in [\mathbf{s}] \ \wedge \ \mathbf{u}(t)= \mathbf{u} \}$.
\end{definition}

\begin{example}\label{ex:symb_state}
For the ACAS Xu, the symbolic state $([\mathbf{s}],u)$ with $[\mathbf{s}]=[-20ft,0ft]$ $\times[8000ft,8500ft]\times[3.10,3.14]\times[700ft/s,700ft/s]\times[600ft/s,600ft/s]$ and $u = 0.0deg/s$ represents a (infinite) set of states where the intruder is ahead of ownship, moving towards the ownship, and the ACAS Xu controller advises COC.
\end{example}

\begin{definition}
A \emph{symbolic set} is a collection of symbolic states defined by $\widetilde{\mathbf{\Phi}} = \{ ([\mathbf{s}]_k,\mathbf{u}_k) \}_{1 \leq k \leq K}$ wherein $K \in \mathbb{N}$. It corresponds to the union of the sets represented by each $([\mathbf{s}]_k,\mathbf{u}_k)$.
\end{definition}

As one can note, a symbolic set can be used to symbolically approximate \emph{any} set of (non-bottom) states of $\mathcal{C}$ (the bottom element is not considered as it does not impact safety). Moreover, our definition yields a rather accurate approximation as it captures the dependency between the state $\mathbf{s}(t)$ of the plant $\mathcal{P}$ and the actuation command $\mathbf{u}(t)$ from the controller. This is made possible as $\mathbf{u}(t)$ can only take a finite number of values.

In the following, we extend the set operations and relations to both symbolic states and symbolic sets \eg $\phi \in \widetilde{\mathbf{\Phi}}$ iff $\phi$ belongs to the set represented by $\widetilde{\mathbf{\Phi}}$. 

\subsection{Over-approximation techniques}\label{sec:approx_techniques}

Our approach for constructing $\widetilde{\mathbf{R}}_{[0,\tau]}$ is to leverage existing over-approximation techniques. More precisely, we aim at using \emph{validated simulation} to soundly approximate the dynamics of the plant $\mathcal{P}$ and \emph{abstract interpretation} to soundly approximate the behaviour of the controller $\mathcal{N}$. These two techniques are presented below and section \ref{sec:main_algo} details how they are combined together to compute $\widetilde{\mathbf{R}}_{[0,\tau]}$. 

\paragraph{Validated simulation} Let us consider an ODE $\mathbf{s}'(t) = f(t,\mathbf{s}(t),\mathbf{u}(t))$ wherein $\mathbf{s}: \mathbb{R} \rightarrow \mathbb{R}^l$, $\mathbf{u}: \mathbb{R} \rightarrow \mathbb{R}^d$ is a given function, continous in $t$, and $f:\mathbb{R} \times \mathbb{R}^l \times \mathbb{R}^d \rightarrow \mathbb{R}^l$ is assumed to be continuous in $t$ and $\mathbf{u}$ and uniformly Lipschitz continuous in $\mathbf{s}$. Moreover, let us consider an interval $[t_1,t_2]$ and a $l$-dimensional box $[\mathbf{s}_{t=t_1}] \subset \mathbb{R}^l$ representing a set of initial values. The goal of validated simulation is to over-approximate the reachable solutions of the ODE satisfying $\mathbf{s}(t=t_1) \in [\mathbf{s}_{t=t_1}]$, over the whole time interval $[t_1,t_2]$. More precisely, it aims at computing the $l$-box $[\mathbf{s}_{[t_1,t_2]}] \subset \mathbb{R}^l$ approximating the reachable values of $\mathbf{s}(t)$ for $t \in [t_1,t_2]$, and the tighter $l$-box $[\mathbf{s}_{t=t_2}] \subset [\mathbf{s}_{[t_1,t_2]}]$ approximating the reachable values of $\mathbf{s}(t)$ at $t=t_2$. Consequently, if $\mathbf{s}$ satisfies the ODE and the initial condition $\mathbf{s}(t=t_1) \in [\mathbf{s}_{t=t_1}]$ then $\left(\mathbf{s}(t) \in [\mathbf{s}_{[t_1,t_2]}] \ \forall t \in [t_1,t_2]\right) \wedge \left(\mathbf{s}(t=t_2) \in [\mathbf{s}_{t=t_2}]\right)$. Usually, validated simulation is based on the $2$-step Löhner type algorithm: the enclosure $[\mathbf{s}_{t_1,t_2]}]$ is calculated using the Banach fixed point theorem while the enclosure $[\mathbf{s}_{t=t_2}]$ is computed based on a numerical integration method (\eg Euler, Runge-Kutta) and the associated local truncation error \cite{sandretto_2015}.

\paragraph{Abstract interpretation} 
Let us consider a function $F:\mathbb{R}^m \rightarrow\mathbb{R}^p$ and let $[\mathbf{x}] \subset \mathbb{R}^m$ be a $m$-dimensional box representing a set of inputs. The goal of abstract interpretation is to soundly approximate the set of the reachable outputs from $[\mathbf{x}]$ \ie the set $F([\mathbf{x}]) = \left\{F(\mathbf{x}) \ | \ \mathbf{x} \in [\mathbf{x}]\right\}$. To this end, abstract interpretation leverages an abstract transformer $F^\#$ that soundly approximates the semantics of $F$. Intuitively, it ``propagates" $[\mathbf{x}]$ through the function $F$. This yields the $p$-box $[\mathbf{y}]  = F^\#([\mathbf{x}])$ satisfying $[\mathbf{y}] \supset F(\mathbf{X})$. The abstract transformer $F^\#$ can rely on interval arithmetics or affine arithmetics for example \cite{stolfi_2003}. 

\subsection{Procedure}\label{sec:main_algo}

In the following, we consider that $\tau$ comprises $q$ executions of the controller \ie $\tau=qT$. The overall idea of our approach is to \emph{iteratively} build the set $\widetilde{\mathbf{R}}_{[0,\tau]}$, based on the successive executions of the controller. To this end, we define a procedure that involves two types of sets:

\begin{enumerate}[label=(\alph*)]
\item The symbolic set \Rtildej{} $\supset \mathbf{R}_{jT} \ \backslash \ \{\bot\}$ approximates the (non-bottom) reachable states at $t=jT$, with $j \leq q$. The $k^{th}$ symbolic state composing \Rtildej{} is denoted \symbstatej{}. It represents a set of states $\mathbf{s}(t)$ that are reachable together with the command $\mathbf{u}_{j,k}$ at $t=jT$.
\item The symbolic set \Rtildestepj{} $\supset \mathbf{R}_{[jT,(j+1)T[} \ \backslash \ \{\bot\}$ approximates the (non-bottom) reachable states for $t \in [jT, (j+1)T[$, with $j < q$. The $k^{th}$ symbolic state composing \Rtildestepj{} is denoted \symbstatestepj{}. It represents a set of states $\mathbf{s}(t)$ that are reachable together with the command $\mathbf{u}_{j,k}$ for $t \in [jT,(j+1)T[$.
\end{enumerate}

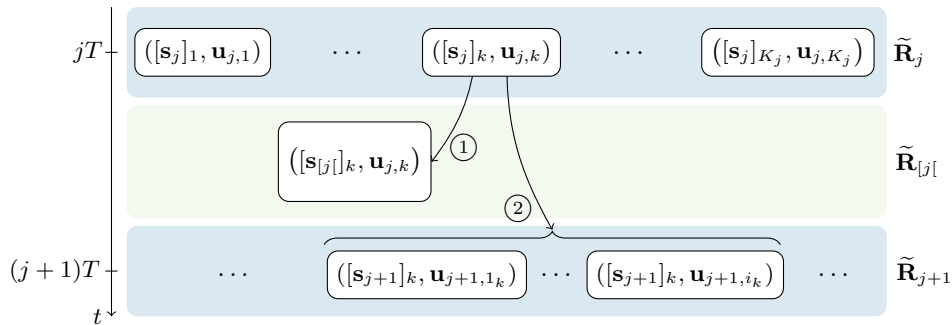
\begin{figure}[hbt]
   \begin{center}
     \begin{tikzpicture}[pnt node/.style={circle,inner sep=0pt, minimum size=0pt, fill=red}]
     \draw[thin,draw=none, rounded corners, fill=MidnightBlue!10] (0,0) rectangle (10,1.2);
     \draw[thin,draw=none, rounded corners, fill=MidnightBlue!10] (0,2.9) rectangle (10,4.1);
     \draw[thin,draw=none, rounded corners, fill=YellowGreen!10] (0,1.3) rectangle (10,2.8);

      \draw [->, bend right = 15] (5,3.5) to node[pos=0.9,left] {\textcircled{\scriptsize{2}}} (5.6,1.15);
      \draw [decorate,decoration={brace,amplitude=5pt}]
  (2.6,0.95) -- (8.6,0.95);
  \node[rectangle, draw=black, rounded corners, fill=white, minimum height=18pt](n5)  at (3.95,0.55) {\small{$\left([\mathbf{s}_{j+1}]_k,\mathbf{u}_{j+1,1_k}\right)$}};
  \node[draw=none,fill=none] (n6) at (5.65,0.55){$\ldots$};
   \node[rectangle, draw=black, rounded corners, fill=white, minimum height=18pt](n7)  at (7.35,0.55) {\small{$\left([\mathbf{s}_{j+1}]_k,\mathbf{u}_{j+1,i_k}\right)$}};
   \node[draw=none,fill=none] (n8) at (1.4,0.55){$\ldots$};
   \node[draw=none,fill=none] (n9) at (9.3,0.55){$\ldots$};
   
   \node[rectangle, draw=black, rounded corners, fill=white, minimum height=30pt](n10)  at (3,2.05) {\small{$\left([\mathbf{s}_{[j[}]_k,\mathbf{u}_{j,k}\right)$}};
   \draw [->, bend left = 15] (4.6,3.5) to node[pos=0.9,right] {\textcircled{\scriptsize{1}}} (n10.east);
   
   \node[rectangle, draw=black, rounded corners, fill=white, minimum height=18pt](n0)  at (1,3.5) {\small{$\left([\mathbf{s}_j]_1,\mathbf{u}_{j,1}\right)$}};
     \node[rectangle, draw=black, rounded corners, fill=white, minimum height=18pt](n1) at (4.8,3.5) {\small{$\left([\mathbf{s}_j]_{k},\mathbf{u}_{j,k}\right)$}};
     \node[rectangle, draw=black, rounded corners, fill=white, minimum height=18pt](n2)  at (8.7,3.5) {\small{$\left([\mathbf{s}_j]_{K_j},\mathbf{u}_{j,K_j}\right)$}};
     \node[draw=none,fill=none] (n3) at (2.9,3.5){$\ldots$};
     \node[draw=none,fill=none] (n4) at (6.6,3.5){$\ldots$};
     
     \draw[->] (-0.2,4.1) -- (-0.2,0) node[left,pos=1]{\small{$t$}};
     \draw[-] (-0.28,0.6) -- (-0.12,0.6) node[left=3pt]{\small{$(j+1)T$}};
     \draw[-] (-0.28,3.5) -- (-0.12,3.5) node[left=2pt]{\small{$jT$}};
     \node (t0) at (10.3,3.5){\small{\Rtildej{}}};
     \node (t1) at (10.5,0.6){\small{\Rtildejnext{}}};
     \node (t2) at (10.4,2.05){\small{\Rtildestepj{}}};

     \end{tikzpicture}
   \end{center}
 \caption{The reachability procedure at control step $j$, where \textcircled{\scriptsize{1}} involves validated simulation and \textcircled{\scriptsize{2}} involves both validated simulation and abstract interpretation.}\label{fig:procedure}
\end{figure}

\noindent The procedure starts with the symbolic set $\widetilde{\mathbf{R}}_0 \supset \mathbf{R}_0 = \mathbf{I}$ enclosing the possible initial states. Then, for $j \in [\![0,q-1]\!]$, it computes the reachable symbolic states from each symbolic state \symbstatej{} composing \Rtildej{} (see Fig. \ref{fig:procedure}). More precisely, for each \symbstatej{} $\in$ \Rtildej{}, it computes:
\begin{enumerate}[label=(\arabic*)]
\setlength\itemsep{1em}
\item The symbolic state \symbstatestepj{} approximating the reachable states from \symbstatej{} over $[jT,(j+1)T[$, where \symbsstepj{} is calculated using validated simulation and $\mathbf{u}_{j,k}$ is the constant actuation command over $[jT,(j+1)T[$. More specifically, to compute \symbsstepj{}, we consider the ODE $\mathbf{s}'(t) = f(t, \mathbf{s}(t), \mathbf{u}(t))$ and the time interval $[jT,(j+1)T]$, with $\mathbf{s}(t=jT) \in$ \symbsj{} and $\mathbf{u}(t) = \mathbf{u}_{j,k} \ \forall t \in [jT,(j+1)T]$. Validated simulation is used to compute the $l$-box $[\mathbf{s}_{[jT,(j+1)T]}]$ enclosing the reachable values of $\mathbf{s}(t)$ for $t \in [jT,(j+1)T]$. Then we take \symbsstepj{} $= [\mathbf{s}_{[jT,(j+1)T]}]$ which is sound as $[jT,(j+1)T[ \subset [jT,(j+1)T]$.
\item The symbolic states $\left([\mathbf{s}_{j+1}]_k,\mathbf{u}_{j+1,1_k}\right),\ldots,\left([\mathbf{s}_{j+1}]_k,\mathbf{u}_{j+1,i_k}\right)$ approximating the reachable states from \symbstatej{} at $t=(j+1)T$, where \symbsjnext{} is calculated using validated simulation and the reachable commands $\mathbf{u}_{j+1,1_k},$ $\ldots,$ $\mathbf{u}_{j+1,i_k}$ are caluclated using abstract interpretation. More specifically, to compute \symbsjnext{}, we consider the same hypotheses as in (1) except that validated simulation is used to compute the $l$-box $[\mathbf{s}_{t=(j+1)T}]$ enclosing the reachable values of $\mathbf{s}(t)$ at $t=(j+1)T$. Then we take \symbsjnext{} $= [\mathbf{s}_{t=(j+1)T}]$ which is sound even though the actuation command may have changed at $t=(j+1)T$ (this is due to the continuity of $\mathbf{s}$). Additionally, to compute the reachable commands, we approximate the behaviour of the controller as follows. First, the network to be executed $N_{j,k}$ is selected based on the previous command \ie $N_{j,k} = \lambda(\mathbf{u}_{j,k})$. Then, abstract interpretation is used to compute: (i) the $m$-box $[\mathbf{x}_{j}]_k = \text{Pre}^\#([\mathbf{s}_j]_k)$ approximating the reachable inputs of the network, (ii) the $p$-box $[\mathbf{y}_j]_k = F_{j,k}^\#([\mathbf{x}_{j}]_k)$ approximating the reachable outputs of the network, where $F_{j,k}$ denotes the function computed by the network $N_{j,k}$ and (iii) the finite set $\{\mathbf{u}_{j+1,1_k},\ldots,\mathbf{u}_{j+1,i_k}\} = \text{Post}^\#([\mathbf{y}_j])$ approximating the reachable commands at $t=(j+1)T$.
\end{enumerate}

By definition, the stages (1) and (2) yield the symbolic sets \Rtildestepj{} and \Rtildejnext{}, the latter being used in the next iteration. Finally, the $q^{th}$ iteration yields $\widetilde{\mathbf{R}}_{[0,\tau]} = \cup_{0 \leq j < q} \widetilde{\mathbf{R}}_{[j[} \ \cup \ \widetilde{\mathbf{R}}_q$ (to be totally rigorous, the bottom element shall be added but this is useless since it does not impact safety).

\medskip

\noindent Actually, to take account of a potential termination of $\mathcal{C}$, we consider a slight variant of the above procedure. Indeed, if a symbolic state \symbstatej{} composing \Rtildej{} satisfies \symbstatej{} $\subset \mathbf{T}$, then this symbolic state is not further propagated \ie the reachable symbolic states from \symbstatej{} are not computed. Consequently, if there exists $j^{\text{end}} \leq q$ such that there is no more symbolic state to be propagated from $\widetilde{\mathbf{R}}_{j_{\text{end}}}$, then we take $\widetilde{\mathbf{R}}_{[0,\tau]} =\cup_{0 \leq j < j^{\text{end}}} \widetilde{\mathbf{R}}_{[j[} \ \cup \ \widetilde{\mathbf{R}}_{j^{\text{end}}}$. Moreover, if $\widetilde{\mathbf{R}}_{[0,\tau]}$ satisfies $\widetilde{\mathbf{R}}_{[0,\tau]} \cap \mathbf{E} = \emptyset$, then the closed-loop $\mathcal{C}$ is proved to be safe \emph{until it terminates}.

\begin{remark}
This mechanism can detect the termination of $\mathcal{C}$ only at the instants $t=T,2T,3T,\ldots$, meaning that the true instants when $\mathcal{C}$ terminates are very likely to be missed. However, this remains a good mechanism when $\mathbf{T}$ behaves like an attractor \ie when $\mathcal{C}$ reaches a state in $\mathbf{T}$ without terminating, then its state tends to stay in $\mathbf{T}$.   
\end{remark}

\begin{theorem}
The procedure yields a sound approximation of the non-bottom reachable states \ie $\widetilde{\mathbf{R}}_{[0,\tau]} \supset \mathbf{R}_{[0,\tau]} \ \backslash \ \{ \bot \}$.
\end{theorem}

\if\short0
\begin{proof} The proof is two fold. 

\begin{enumerate}[label=(\roman*),itemsep=1pt,topsep=1pt]
\item First, let us show by induction that \Rtildej{} soundly approximates the non-bottom reachable states at $t=jT$ \ie \Rtildej{} $\supset \mathbf{R}_{jT} \ \backslash \ \{\bot\}$ for $0 \leq j \leq q$. By definition, $\widetilde{\mathbf{R}}_0 \supset \mathbf{R}_0 = \mathbf{I}$ is a sound approximation of the non-bottom reachable states at $t=0$. Let $\phi^* \neq \bot$ be a reachable state at $t^*=(j+1)T$. Given the definition of a reachable state (see definition \ref{def:Rt}), there exists a unique function $\phi^*_{\phi_0}: [0,\tau] \rightarrow \mathbb{R}^l \times \mathbf{U} \cup \{\bot\}$ such that $\phi^* = \phi^*_{\phi_0}((j+1)T)$. Moreover, since $\phi^* \neq \bot$, neither the target set $\mathbf{T}$ nor the bottom state have been reached already \ie $\phi^*_{\phi_0}(t) \notin \mathbf{T} \cup \{\bot\} \  \forall t < (j+1)T$. By induction, there is a symbolic state \symbstatej{} $\in$ \Rtildej{} such that $\phi^*_{\phi_0}(jT) \in$ \symbstatej{}. Additionally, as $\phi^*_{\phi_0}(jT) \notin \mathbf{T}$,  \symbstatej{} satisfies \symbstatej{} $\not\subset \mathbf{T}$. Consequently, the procedure computes the reachable states from \symbstatej{} at $t=(j+1)T$, yielding the symbolic states $\left([\mathbf{s}_{j+1}]_k,\mathbf{u}_{j+1,1_k}\right),\ldots,\left([\mathbf{s}_{j+1}]_k,\mathbf{u}_{j+1,i_k}\right)$. As validated simulation and abstract interpretation are sound, these symbolic states constitute a sound approximation. Therefore, $\phi^*_{\phi_0}((j+1)T) \in \left([\mathbf{s}_{j+1}]_k,\mathbf{u}_{j+1,1_k}\right),\ldots,\left([\mathbf{s}_{j+1}]_k,\mathbf{u}_{j+1,i_k}\right) \subset$ \Rtildejnext{}. Hence $\phi^* \in$ \Rtildejnext{}.

\item Secondly, let us show that \Rtildestepj{} soundly approximates the non-bottom reachable states over the time interval $[jT,(j+1)T[$ \ie \Rtildestepj{} $\supset \mathbf{R}_{[jT,(j+1)T[} \ \backslash \ \{\bot\}$ for $0 \leq j < q$. Let $\phi^* \neq \bot$ be a reachable state at $t^* \in [jT,(j+1)T[$. There exists a unique function $\phi^*_{\phi_0}$ such that $\phi^* = \phi^*_{\phi_0}(t^*)$ and  $\phi^*_{\phi_0}(t) \notin \mathbf{T} \cup \{\bot\} \  \forall t < t^*$. As shown before, $\phi^*_{\phi_0}(jT) \in$ \Rtildej{}, so there is a symbolic state \symbstatej{} $\in$ \Rtildej{} such that $\phi^*_{\phi_0}(jT) \in$ \symbstatej{}. Additionally, since $\phi^*_{\phi_0}(jT) \notin \mathbf{T}$,  \symbstatej{} satisfies \symbstatej{} $\not\subset \mathbf{T}$. Consequently, the procedure computes the reachable states from \symbstatej{} over the interval $[jT,(j+1)T[$, yielding the symbolic state \symbstatestepj{}. As validated simulation is sound and the command signal remains constant over the interval $[jT,(j+1)T[$, this symbolic state constitutes a sound approximation. Therefore, $\phi^*_{\phi_0}(t^*) \in$ \symbstatestepj{} $\subset$ \Rtildestepj{}. Hence $\phi^* \in$ \Rtildestepj{}.

\end{enumerate}
\hspace*{\fill}$\square$

\end{proof}
\fi

\if\short0
\subsection{Optimizations}

\paragraph{Improving precision} In the above procedure, a \emph{single} $l$-box \symbsstepj{} encloses the reachable states $\mathbf{s}(t)$ from the symbolic state \symbstatej{} over $[jT,(j+1)T[$. Due to its shape, the $l$-box \symbsstepj{} may contain a lot of unreachable states, resulting in a loose approximation (see Fig. \ref{fig:validated_sim}). In order to yield a tighter approximation, the procedure is slightly modified. Instead of using a single $l$-box to approximate the reachable states $\mathbf{s}(t)$, a collection of $M > 1$ $l$-boxes is used. This collection of $l$-boxes is obtained by performing $M$ integration steps \ie $M$ successive validated simulations. More precisely, we start with the $l$-box $[\mathbf{s}_{j,0}]_k \triangleq$ \symbsj{}. Then, for $i \in [\![0,M-1]\!]$, we consider the ODE $\mathbf{s}'(t) = f(t,\mathbf{s}(t),\mathbf{u}(t))$ and the time interval $\left[\left(j + \frac{i}{M}\right)T,\left(j + \frac{i+1}{M}\right)T\right]$, with $\mathbf{s}\left(t= \left(j + \frac{i}{M}\right)T\right) \in [\mathbf{s}_{j,i}]_k$ and $\mathbf{u}(t) = \mathbf{u}_{j,k} \ \forall t \in \left[\left(j + \frac{i}{M}\right)T,\left(j + \frac{i+1}{M}\right)T\right]$. Validated simulation is used to compute (1) the $l$-box $[\mathbf{s}_{[j[,i}]_k$ approximating the reachable states $\mathbf{s}(t)$ over $\left[\left(j + \frac{i}{M}\right)T,\left(j + \frac{i+1}{M}\right)T\right[$ and (2) the $l$-box $[\mathbf{s}_{j,i+1}]_k$ enclosing the reachable states $\mathbf{s}(t)$ at $t=\left(j + \frac{i+1}{M}\right)T$. The latter $l$-box is used to perform next integration step. Finally, we take \symbsstepj{} $= \left\{ [\mathbf{s}_{[j[,i}]_k \right\}_{0 \leq i < M}$ and also \symbsjnext{} $= [\mathbf{s}_{j,M}]_k$.

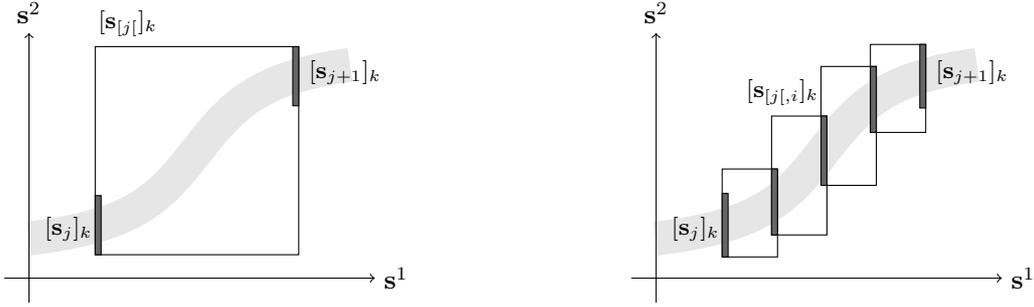
\begin{figure}[hbt]
\begin{minipage}[c]{0.45\linewidth}
\centering
\def\radius{0.45}
\def\step{2}
\def\tbegin{0.7}
\def\tend{2.7}
\def\w{0.03}
\def\factor{1.3}
\begin{tikzpicture}[scale=1.3][x=1cm,y=1cm]
\draw[scale=1, domain=0:3.25, smooth, variable=\x, black!10, line width = \radius cm] plot ({\x}, {rad(atan((\x-1.7)*1.8))/1.4 + 1.3});
\draw[->] (-0.25,0) -- (3.5,0) node[right] {$\mathbf{s}^1$};
\draw[->] (0,-0.25) -- (0,2.5) node[above] {$\mathbf{s}^2$};

\foreach \z in {0.7}
	\draw (\z - \w, {rad(atan((\z - \w -1.7)*1.8))/1.4 + 1.3 - \radius/2 * \factor}) rectangle (\z + \step + \w, {rad(atan((\z + \step + \w -1.7)*1.8))/1.4 + 1.3 + \radius/2 * \factor});
	
\foreach \z in {0.7}
\draw[fill=black!60] (\z + \step - \w, {rad(atan((\z + \step - \w -1.7)*1.8))/1.4 + 1.3 - \radius/2 * \factor}) rectangle (\z + \step + \w, {rad(atan((\z + \step + \w -1.7)*1.8))/1.4 + 1.3 + \radius/2 * \factor});
	
\draw[fill=black!60] (\tbegin - \w, {rad(atan((\tbegin - \w -1.7)*1.8))/1.4 + 1.3 - \radius/2 * \factor}) rectangle (\tbegin + \w, {rad(atan((\tbegin + \w -1.7)*1.8))/1.4 + 1.3 + \radius/2 * \factor});

\node (n0) at ({\tbegin - 0.3}, 0.5) {\small{$[\mathbf{s}_j]_k$}};
\node (n1) at ({\tend +0.5}, 2.1) {\small{$[\mathbf{s}_{j+1}]_k$}};
\node (n2) at ({\tbegin + 0.3}, 2.6) {\small{$[\mathbf{s}_{[j[}]_k$}};




\end{tikzpicture}
\end{minipage} \hfill
\begin{minipage}[c]{0.45\linewidth}
\centering
\def\radius{0.45}
\def\step{0.5}
\def\tbegin{0.7}
\def\tend{2.7}
\def\w{0.03}
\def\factor{1.4}
\begin{tikzpicture}[scale=1.3][x=1cm,y=1cm]
\draw[scale=1, domain=0:3.25, smooth, variable=\x, black!10, line width = \radius cm] plot ({\x}, {rad(atan((\x-1.7)*1.8))/1.4 + 1.3});
\draw[->] (-0.25,0) -- (3.5,0) node[right] {$\mathbf{s}^1$};
\draw[->] (0,-0.25) -- (0,2.5) node[above] {$\mathbf{s}^2$};

\foreach \z in {0.7,1.2,1.7,2.2}
	\draw (\z - \w, {rad(atan((\z - \w -1.7)*1.8))/1.4 + 1.3 - \radius/2 * \factor}) rectangle (\z + \step + \w, {rad(atan((\z + \step + \w -1.7)*1.8))/1.4 + 1.3 + \radius/2 * \factor});
	
\foreach \z in {0.7,1.2,1.7,2.2}
\draw[fill=black!60] (\z + \step - \w, {rad(atan((\z + \step - \w -1.7)*1.8))/1.4 + 1.3 - \radius/2 * \factor}) rectangle (\z + \step + \w, {rad(atan((\z + \step + \w -1.7)*1.8))/1.4 + 1.3 + \radius/2 * \factor});
	
\draw[fill=black!60] (\tbegin - \w, {rad(atan((\tbegin - \w -1.7)*1.8))/1.4 + 1.3 - \radius/2 * \factor}) rectangle (\tbegin + \w, {rad(atan((\tbegin + \w -1.7)*1.8))/1.4 + 1.3 + \radius/2 * \factor});

\node (n0) at ({\tbegin - 0.3}, 0.5) {\small{$[\mathbf{s}_j]_k$}};
\node (n1) at ({\tend +0.5}, 2.1) {\small{$[\mathbf{s}_{j+1}]_k$}};
\node (n2) at (1.28, 1.9) {\small{$[\mathbf{s}_{[j[,i}]_k$}};




%

\end{tikzpicture}
\end{minipage}
\caption{Over-approximation of the plant dynamics using validated simulation with a single integration step (left) and $M=4$ integration steps (right).}
\label{fig:validated_sim}
\end{figure}

\setlength{\commentindent}{.25\textwidth}
\renewcommand{\algorithmicrequire}{\textbf{Input:}}
\renewcommand{\algorithmicensure}{\textbf{Output:}}
\begin{algorithm}
\caption{Over-approximation of the plant dynamics.}
\label{alg:simulating}
\begin{algorithmic}[1]
\Require The function $f$ describing the plant dynamics, the execution period $T$ of the controller, the number of integration steps $M$ and the symbolic state \symbstatej{}.
\Ensure The $l$-boxes \symbsstepj{} and \symbsjnext{}.
\Function{Simulate}{$f$, $T$, $M$, \symbstatej{}}
\State $[\mathbf{s}_{j,0}]_k$ $\leftarrow$ \symbsj{} 
\For{$i \in [\![0,M-1]\!]$}
\State $\left( [\mathbf{s}_{[j[,i}]_k, [\mathbf{s}_{j,i+1}]_k\right)$ $\leftarrow$ validatedSimulation$\left(f,\left[\left(j + \frac{i}{M}\right)T,\left(j + \frac{i+1}{M}\right)T\right],[\mathbf{s}_{j,i}]_k,\mathbf{u}_{j,k}\right)$
\EndFor
\State \symbsstepj{} $\leftarrow$ $\left\{ [\mathbf{s}_{[j[,0}]_k, \ldots,  [\mathbf{s}_{[j[,M-1}]_k\right\}$
\State \symbsjnext{} $\leftarrow$ $[\mathbf{s}_{j,M}]_k$
\State \Return $\left( \text{\symbsstepj{}}, \text{\symbsjnext{}} \right)$
\EndFunction
\end{algorithmic}
\end{algorithm}

\paragraph{Improving time complexity} In the worst case, the number of symbolic states in \Rtildej{} grows exponentially with $j$. Indeed, each symbolic state \symbstatej{} composing \Rtildej{} can lead up to $P$ symbolic states $\left([\mathbf{s}_{j+1}]_k,\mathbf{u}_{j+1,1_k}\right),\ldots,\left([\mathbf{s}_{j+1}]_k,\mathbf{u}_{j+1,i_k}\right)$ in \Rtildejnext{} (recall that $P$ is the number of elements in $\mathbf{U}$ \ie the number of possible actuation commands). In order to avoid an exponential blow up, the procedure is slightly modified by keeping the number of symbolic states in \Rtildej{} below a given threshold $\Gamma$, for all $j \in [\![0,q]\!]$. As a result, provided we can bound the execution time of validated simulation and abstract interpretation, the time complexity of the procedure remains linear with $q$. For keeping the size of \Rtildej{} below $\Gamma$, some symbolic states are \emph{joined} based on a dedicated heuristics. This heuristics uses the notion of \emph{distance} between two symbolic states as well as a \emph{join} operation. 

\begin{definition}\label{def:distance}
The \emph{distance} between two symbolic states $\left([\mathbf{s}]_1,\mathbf{u}\right)$ and $\left([\mathbf{s}]_2,\mathbf{u}\right)$ with \emph{same} actuation command $\mathbf{u}$ is defined as the euclidean distance between the centers of the $l$-boxes $[\mathbf{s}]_1$ and $[\mathbf{s}]_2$:
\begin{equation}
d\left( \left([\mathbf{s}]_1,\mathbf{u}\right), \left([\mathbf{s}]_2,\mathbf{u}\right) \right) = || C_1 - C_2 ||_2^2
\end{equation}
wherein $C_1$ (resp $C_2$) is a $l$-dimensional vector of which the $i^{th}$ component is the center of the $i^{th}$ interval composing $[\mathbf{s}]_1$ (resp $[\mathbf{s}]_2$).
\end{definition}

\begin{definition}\label{def:join_operation}
The \emph{join} operation takes as inputs two symbolic states $\left([\mathbf{s}]_1,\mathbf{u}\right)$ and $\left([\mathbf{s}]_2,\mathbf{u}\right)$ with \emph{same} actuation command $\mathbf{u}$ and outputs a symbolic state $\left([\mathbf{s}]_3,\mathbf{u}\right)$ such that $[\mathbf{s}]_3$ is the smaller $l$-box containing both $[\mathbf{s}]_1$ and $[\mathbf{s}]_2$.
\end{definition}

The heuristics works as follows. At the $j^{th}$ control step, if the number of symbolic states $K_j$ in \Rtildej{} is greater than $\Gamma$, then the symbolic states composing \Rtildej{} are clustered into $P$ groups, each one corresponding to a given actuation command. More specifically, the $i^{th}$ group is $\mathscr{G}_i = \left\{ \left([\mathbf{s}_j]_k,\mathbf{u}_{j,k}\right) \in \widetilde{\mathbf{R}}_j \ | \ \mathbf{u}_{j,k} = \mathbf{u}^{(i)} \right\}$ where  $\mathbf{u}^{(i)}$ is the $i^{th}$ element of the set $\mathbf{U}$ of the possible actuation commands (see section \ref{sec:closed_loop}). For each group $\mathscr{G}_i$, a distance matrix $\mathscr{D}_i$ is calculated based on definition \ref{def:distance}. Then, the distance matrices $\mathscr{D}_1,\ldots,\mathscr{D}_P$ are used to identify the two \emph{closest} symbolic states in \Rtildej{} (note that these two closest symbolic states necessarily have the same actuation command). Finally, using the join operation introduced in definition \ref{def:join_operation}, the two closest symbolic states are joined. The set \Rtildej{} is updated accordingly and the process is repeated until $K_j \leq \Gamma$ (see Algorithm \ref{alg:clustering}).

The choice of the threshold $\Gamma$ allows a trade-off between accuracy (large $\Gamma$) and computational efficiency (small $\Gamma$).

\setlength{\commentindent}{.18\textwidth}
\renewcommand{\algorithmicrequire}{\textbf{Input:}}
\renewcommand{\algorithmicensure}{\textbf{Ensure:}}
\begin{algorithm}
\caption{Heuristics for keeping the number of symbolic states in \Rtildej{} below the threshold $\Gamma$.}
\label{alg:clustering}
\begin{algorithmic}[1]
\Require The symbolic set \Rtildej{} and the threshold $\Gamma$.
\Ensure length(\Rtildej{}) $\leq \Gamma$ and \Rtildej{} $\supset$ old(\Rtildej{}).
\Procedure{Resize}{\Rtildej{}, $\Gamma$}
\State $K_j$ $\leftarrow$ length(\Rtildej{})
\While{$K_j > \Gamma$}
\For{$i \in [\![1,P]\!]$}
\State $\mathscr{G}_i$ $\leftarrow$ $\left\{ \left([\mathbf{s}_j]_k,\mathbf{u}_{j,k}\right) \in \widetilde{\mathbf{R}}_j \ | \ \mathbf{u}_{j,k} = \mathbf{u}^{(i)} \right\}$
\State $\mathscr{D}_i$ $\leftarrow$ calculateDistanceMatrix($\mathscr{G}_i$)
\EndFor
\State $\left(  \left([\mathbf{s}_j]_{k_1},\mathbf{u}_{j,k_1}\right), \left([\mathbf{s}_j]_{k_2},\mathbf{u}_{j,k_2}\right) \right)$ $\leftarrow$ findClosestStates$\left( \mathscr{D}_1,\ldots,\mathscr{D}_P\right)$
\Comment{$\mathbf{u}_{j,k_1}=\mathbf{u}_{j,k_2}$}
\State $\left([\mathbf{s}_j]_{k_3},\mathbf{u}_{j,k_3}\right)$ $\leftarrow$ join$\left(  \left([\mathbf{s}_j]_{k_1},\mathbf{u}_{j,k_1}\right), \left([\mathbf{s}_j]_{k_2},\mathbf{u}_{j,k_2}\right) \right)$
\Comment{$\mathbf{u}_{j,k_3}=\mathbf{u}_{j,k_1}$}
\State \Rtildej{} $\leftarrow$ $\left(\widetilde{\mathbf{R}}_j \ \backslash \ \left\{ \left([\mathbf{s}_j]_{k_1},\mathbf{u}_{j,k_1}\right), \left([\mathbf{s}_j]_{k_2},\mathbf{u}_{j,k_2}\right) \right\} \right) \cup \left\{\left([\mathbf{s}_j]_{k_3},\mathbf{u}_{j,k_3}\right)\right\}$
\State $K_j$ $\leftarrow$ $K_j - 1$
\EndWhile
\EndProcedure
\end{algorithmic}
\end{algorithm}

\begin{remark}
As one may note, $\Gamma$ must be chosen greater than $P$. Indeed, two symbolic states with two different actuation commands cannot be joined so the heuristics would fail to keep the number of symbolic states in \Rtildej{} strictly below $P$. 
\end{remark}

\subsection{Overall algorithm}

\setlength{\commentindent}{.39\textwidth}
\renewcommand{\algorithmicrequire}{\textbf{Input:}}
\renewcommand{\algorithmicensure}{\textbf{Output:}}
\begin{algorithm}[H]
\caption{Reachability analysis of the closed-loop system $\mathcal{C}$.}
\label{alg:simu}
\begin{algorithmic}[1]
\Require The closed-loop system $\mathcal{C} = \left(\mathcal{P}, \mathcal{N}\right)$, the approximated set of the initial states $\widetilde{\mathbf{R}}_0 \supseteq \mathbf{R}_0 = \mathbf{I}$, the set of the erroneous states $\mathbf{E}$, the target set $\mathbf{T}$, the number of control steps $q$, the number of integration steps $M$ and the threshold $\Gamma$.
\Ensure A Boolean indicating whether the closed-loop $\mathcal{C}$ is proved safe until it terminates.

\State $j_{\text{end}}$ $\leftarrow$ $q$
\State hasTerminated $\leftarrow$ False
\For{$j \in [\![0,q-1]\!]$}
\State $\triangleright$ keep the size of \Rtildej{} below $\Gamma$ (see Algorithm \ref{alg:clustering})
\State  \textsc{Resize}(\Rtildej{},$\Gamma$)
\State $\triangleright$ compute the symbolic sets \Rtildestepj{} and \Rtildejnext{}
\State \Rtildestepj{} $\leftarrow$ $\emptyset$
\State \Rtildejnext{} $\leftarrow$ $\emptyset$
\For{$\left( \left([\mathbf{s}_j]_k,\mathbf{u}_{j,k}\right) \in \widetilde{\mathbf{R}}_j \ \wedge \ \left([\mathbf{s}_j]_k,\mathbf{u}_{j,k}\right) \not\subset \mathbf{T} \right)$}
\State $\triangleright$ approximate the dynamics of the plant (see Algorithm \ref{alg:simulating})
\State $\left([\mathbf{s}_{[j[}]_k,[\mathbf{s}_{j+1}]_k\right)$ $\leftarrow$  \textsc{Simulate}$
\left(f,T,M,([\mathbf{s}_j]_k, \mathbf{u}_{j,k})\right)$ 
\State $\triangleright$ approximate the behaviour of the controller 
\State $N_{j,k}$ $\leftarrow$ $\lambda\left(\mathbf{u}_{j,k}\right)$ 
\State $[\mathbf{x}_j]_k$ $\leftarrow$ $\text{Pre}^\#\left([\mathbf{s}_j]_k\right)$
\State $[\mathbf{y}_j]_k$ $\leftarrow$ $F_{j,k}^\#\left([\mathbf{x}_j]_k\right)$
\State $\{\mathbf{u}_{j+1,1_k},\ldots,\mathbf{u}_{j+1,i_k}\}$ $\leftarrow$ $\text{Post}^\#\left([\mathbf{y}_j]_k\right)$
\State $\triangleright$ update the symbolic set \Rtildestepj{} approximating the reachable states over $[jT,(j+1)T[$
\State \Rtildestepj{} $\leftarrow$ \Rtildestepj{} $\cup$ $\{ \text{\symbstatestepj{}} \}$
\State $\triangleright$ update the symbolic set \Rtildejnext{} approximating the reachable states at $t=(j+1)T$
\State \Rtildejnext{} $\leftarrow$  \Rtildejnext{} $\cup$ $\left\{ \left([\mathbf{s}_{j+1}]_k, \mathbf{u}_{j+1,1_k}\right),\ldots,  \left([\mathbf{s}_{j+1}]_k, \mathbf{u}_{j+1,i_k}\right)\right\} $ 
\EndFor
\State $\triangleright$ check for termination
\If{$\widetilde{\mathbf{R}}_{j+1}  \subset \mathbf{T}$}
\State $j_{\text{end}}$ $\leftarrow$ $j+1$
\State hasTerminated $\leftarrow$ True
\State \textbf{break}
\EndIf
\EndFor
\State $\triangleright$ construct the symbolic set $\widetilde{\mathbf{R}}_{[0,\tau]}$
\State $\widetilde{\mathbf{R}}_{[0,\tau]}$ $\leftarrow$ $\widetilde{\mathbf{R}}_{[0[} \cup \ldots \cup\widetilde{\mathbf{R}}_{[j_{\text{end}}-1[} \cup \widetilde{\mathbf{R}}_{j_{\text{end}}} $
\State \Return $\left(\widetilde{\mathbf{R}}_{[0,\tau]} \cap \mathbf{E} = \emptyset \ \wedge \ \text{hasTerminated}\right)$
\end{algorithmic}
\end{algorithm}

\subsection{Implementation details}

We implemented our procedure as a Python program that interfaces with existing tools. The validated simulation of the plant dynamics is based on DynIBEX \cite{sandretto_2015}. The abstract transformers $\text{Pre}^\#$ and $\text{Post}^\#$ approximating the semantics of the pre- and post-processing functions are based on interval arithmetics, which has the advantage to be easy to implement and to offer a computationally efficient analysis while still being accurate for simple functions. The abstract transformer of the neural network function $F_{j,k}^\#$ relies on a dedicated tool named ReluVal \cite{wang_2018}, which uses interval arithmetics together with symbolic interval propagation.

\section{Experiments}\label{sec:experiments}
\subsection{Experimental setup}

\paragraph{Partitioning} For verifying the ACAS Xu, we used an empirical partitioning of the possible initial states. More precisely, the circle $\mathcal{R}$ representing the possible initial positions $(x_0,y_0)$ of the intruder was partitioned into $629$ arcs of length $80 \ \text{ft}$ each. Additionally, for each arc, the possible initial headings $\psi_0$ of the intruder were partitioned into $316$ subsets of size $0.01 \ \text{rad}$ each (see Fig  \ref{fig:initial_set_partition}). With the initial velocities $v_{\text{own},0}$ and $v_{\text{int},0}$ being fixed, we obtained a partition of size $K_0 = 198,764$ of the possible initial states $\mathbf{s}_0$ of the plant $\mathcal{P}$. Then, each element of this partition was over-approximated by a $5$-dimensional box $[\mathbf{s}_0]_k \subset \mathbb{R}^5$, with $1 \leq k \leq K_0$. Finally, we took as input for the procedure the symbolic set $\widetilde{\mathbf{R}}_0 = \{ ([\mathbf{s}_0]_k, 0.0 \ \text{deg/s}) \}_{1 \leq k \leq K_0}$.

\begin{figure}[hbt]
\centering
\pgfplotsset{width=0.45\textwidth,compat=1.8}
\pgfplotsset{
    axis line style={black},
    every tick label/.append style={black}
  }
\begin{tikzpicture}
  \begin{axis}[
    view = {25}{43},
    xlabel = {$x_0$},
    ylabel = {$y_0$},
    zlabel = {$\psi_0$},
    zmin=-5.1,
    zmax=5.1,
    xmin=-1.15,
    xmax=1.15,
    ymin=-1.15,
    ymax=1.15,
    ztick = {-4.712,0,4.712},
    zticklabels = {$-3\pi/2$,$0$,$3\pi/2$},
    xtick={-1,0,1},
    ytick={-1,0,1},
    xticklabels={$-r$,$0$,$r$},
    yticklabels={$-r$,$0$,$r$}
  ]
    \addplot3 [
    samples=50,
    domain=0:2*pi, 
    samples y=0,
    dashed
  ] (
    {cos(deg(x))},
    {sin(deg(x))},
    {-5.1}
  ) node[above, pos=1.07] {$\mathcal{R}$};
  \addplot3 [
    surf,
    colormap/greenyellow,
    shader     = faceted interp,
    point meta = x,
    samples    = 50,
    samples y  = 5,
    z buffer   = sort,
    domain     = 0:2*pi,
    y domain   =-3*pi/2:-pi/2,
    draw = black,
    thick
  ] (
    {-sin(deg(x))},
    {cos(deg(x))},
    {x + y}
  );
  \end{axis}
\end{tikzpicture}
\caption{The ribbon-like set of the possible initial states $\left(x_0,y_0,\psi_0\right)$ with an example of partition represented as a mesh grid.}\label{fig:initial_set_partition}
\end{figure}
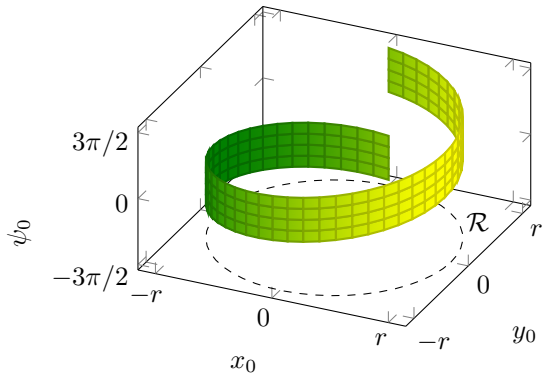

The reason for partitioning was three fold. First, a single initial symbolic state $([\mathbf{s}_0],0.0 \ \text{deg/s})$ approximating $\mathbf{I}$ necessarily contains the unsafe set $\mathbf{E}$, due to the shape of $[\mathbf{s}_0]$ (a box containing the circle $\mathcal{R}$ also contains the center of $\mathcal{R}$ corresponding to a collision between the two aircraft). Secondly, the $K_0$ initial symbolic states composing $\widetilde{\mathbf{R}}_0$ can be seen as $K_0$ \emph{independent} verification problems, of which resolution can be parallelized. Finally, the smaller the box $[\mathbf{s}_0]_k$, the more precise the reachability analysis since the function $f$ representing the dynamics of $\mathcal{P}$ is uniformly Lipschitz continuous in $\mathbf{s}$ and the functions computed by the neural networks are also uniformly Lipschitz continuous \cite{wang_2018}. 

\paragraph{Split refinement} For the same reason as mentioned above, when the system could not be proved safe for a given initial symbolic state $([\mathbf{s}_0]_k,0.0 \ \text{deg/s})$, then this initial symbolic state was splitted into smaller initial symbolic states, leading to a new reachability analysis. More precisely, $[\mathbf{s}_0]_k$ was bisected along the dimensions corresponding to $x_0$, $y_0$ and $\psi_0$, yielding $2^3$ new initial symbolic states. This split refinement process was repeated iteratively until the system could be proved safe, with a maximum depth of $2$.

\medskip

\noindent The experiment was conducted using $M=10$ for the number of integration steps and $\Gamma=P=5$ for the threshold on the number of symbolic states in \Rtildej{}. Moreover, it was run on CentOS 7 with 2 Intel\textsuperscript{\tiny\textregistered} Xeon\textsuperscript{\tiny\textregistered} processors E5-2670 v3 @ 2.30GHz of 12 cores (24 threads) each and 64 GB RAM.

\subsection{Results}

\begin{figure}[hbt]
\begin{minipage}[c]{0.45\linewidth}
\centering
\includegraphics[scale=0.28]{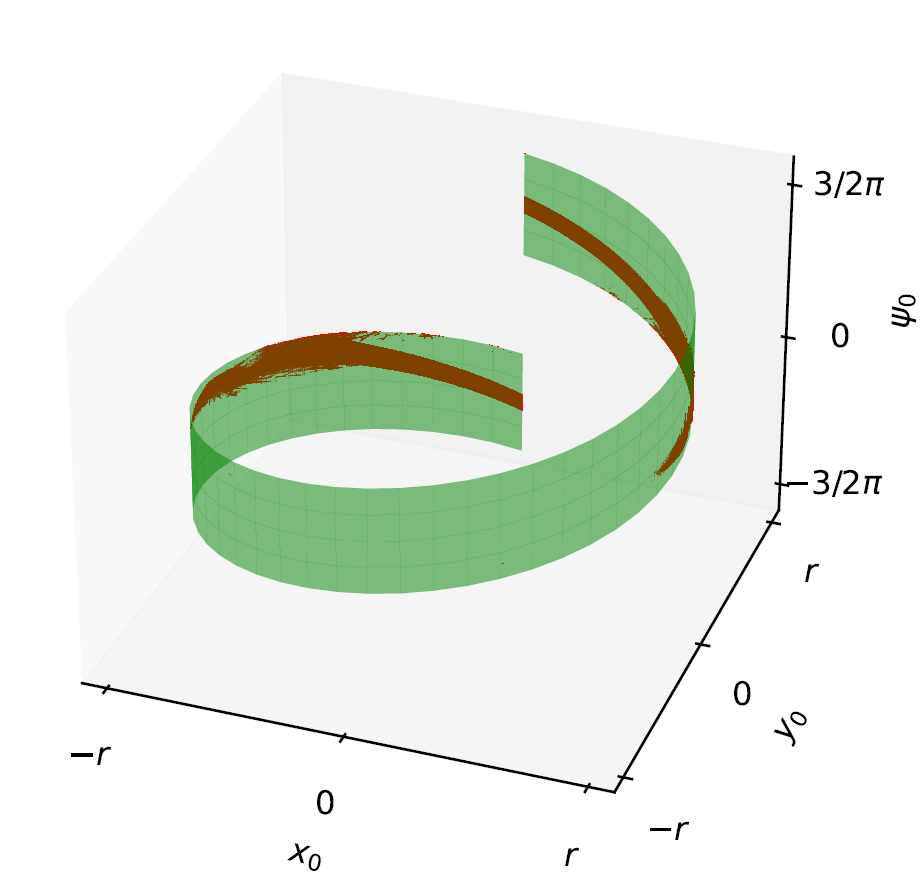}\\
(a)
\end{minipage} \hfill
\begin{minipage}[c]{0.45\linewidth}
\centering
\includegraphics[scale=0.55]{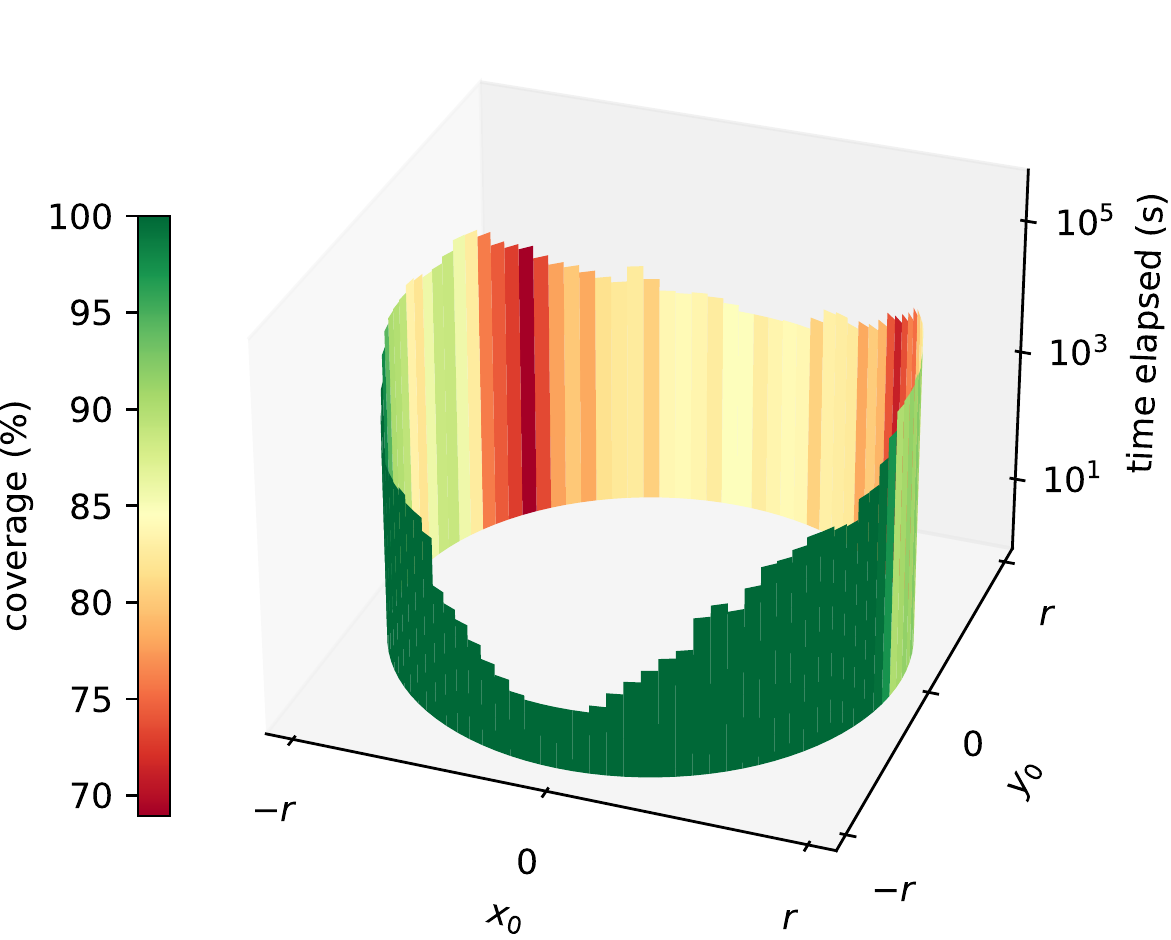}\\
(b)
\end{minipage}
\caption{(a) The initial states for which the system was proved safe (in green) and the initial states for which the system could not be proved safe (in red), (b) The coverage and time elapsed \textit{w.r.t.} the initial position of the intruder (each bar corresponds to a subset of the initial states where the position $(x_0,y_0)$ of the intruder lies along an arc of length $500$ft).}\label{fig:results}
\end{figure}

In our experiment, we recorded (1) the \emph{time elapsed} \ie the time necessary for performing the reachability analysis and (2) the \emph{coverage} $c$ representing the percentage of the possible initial states for which the ACAS Xu was proved safe until it terminates. More precisely, the coverage $c$ was calculated as follows: $c = 100 / K_0 \cdot \sum_{d=0}^{2} n_d / (2^3)^d$ wherein $n_d$ is the number of initial symbolic states resulting from $d$ split refinements and for which the ACAS Xu was proved safe. The reachability analysis took about $12$ days and yielded a coverage $c=90.3$\%, meaning that the ACAS Xu was proved safe for $90.3$\% of the possible initial states.

Although we did not obtain a complete proof of safety, we could leverage the partition of the set $\mathbf{I}$ to identify the initial states for which the ACAS Xu was proved safe and the initial states for which it could not be proved safe (see Fig \ref{fig:results}.a). It is worth noting that such a result represents a valuable information from a practical point of view. For instance, it could be used to design a real-time monitoring mechanism that switches to a more robust controller if the system encounters an initial state for which it was not proved safe. 

The results that we obtained by partitioning the set $\mathbf{I}$ also constitute a valuable information in terms of ``\emph{explainability}", in the sense that they help understanding the behaviour of the overall system. For instance, as one can see in Fig \ref{fig:results}.b, the initial states that yielded the hardest verification tasks correspond to the cases where the intruder is approaching from the left ($x_0<0 \ \wedge \ y_0>0$) or approaching from the right ($x_0>0 \ \wedge \ y_0>0$). Indeed, the coverage obtained in these regions is around $75$\% while it ranges from $85$\% to $100$\% elsewhere. Additionally, in these regions, the time necessary for performing the reachability analysis is about $5\cdot10^4s$ while it is around or below $10^3s$ elsewhere. Such a result provides an interesting information about the potential weaknesses of the controller, which can be interpreted at the system level. It suggests that the most critical situations are encountered not when the intruder is directly ahead of the ownship but when it approaches from the left or from the right. In addition to representing a valuable knowledge about the behaviour of the system, this information could be used to generate new data with the aim of retraining the networks for example. Furthermore, as one can see in Fig. \ref{fig:results}.b, the results are roughly symmetrical \textit{w.r.t.} the $x_0=0$ axis, both in terms of coverage and time elapsed, which suggests that the system has a similar behaviour for two initial states that are symmetrical \textit{w.r.t.} the $x_0=0$ axis. It is worth noting that such a behaviour is quite consistent since the collision avoidance problem is totally symmetrical \textit{w.r.t.} the $x_0=0$ axis. We believe that such an information can help building ``\textit{trust}" in the overall system.

\section{Conclusion and future work}

This paper presented a technique to verify the safety requirements of complex neural network controlled systems such as the ACAS Xu. The proposed technique leverages a generic model of a neural network controlled system together with a reachability analysis, combining validated simulation and abstract interpretation. We evaluated the applicability of our approach by providing the first sound guarantees of safety of the overall neural network based ACAS Xu. Although we could not obtain a complete proof of safety, we showed that our approach can provide valuable information from a practical point of view. 

For future work, instead of using a uniform, empirically-generated partition of the initial states, we aim at finding a more efficient partitioning strategy. For example, we could explore the techniques employed in similar problems such as meshing generation in computational fluid dynamics. Another direction is to propose an efficient heuristics for splitting the initial symbolic states when the system cannot be proved safe. Instead of using a simple bisection along each dimension, we could identify the variable having the most influence on the overall system behaviour, and split along the corresponding dimension only. A third direction is to combine our approach with an efficient falsification strategy that can search for unsafe trajectories when the system cannot be proved safe. Lastly, for the ACAS Xu, we could consider multiple UAVs, each one being equipped with a collision avoidance controller. Indeed, our model and procedure only have to be slightly adapted to represent multiple agents interacting together, all equipped with a controller. The plant could capture the dynamics of the multiple agents (the same way we captured the dynamics of both the ownship and the intruder) and be combined with several controllers. Then, instead of evaluating one controller, our procedure would evaluate several controllers, which is straightforward if all the controllers execute in the same time interval.

\bibliographystyle{plain}
\bibliography{references}

\end{document}